\documentclass[letterpaper]{article} 
\newcommand{\ours}{RealTOD}
\newcommand{\stitle}[1]{\noindent\textup{\textbf{#1}}}
\newcommand{\myNum}[1]{(\emph{#1})}
\newcommand{\cmark}{\textcolor{green}{\ding{51}}}  
\newcommand{\xmark}{\textcolor{red}{\ding{55}}}    

\newcommand{\LlamaLongName}{Llama-3.3-70B-Instruct}
\newcommand{\LlamaShortName}{Llama}

\newcommand{\DeepSeekLongName}{DeepSeek-V3}
\newcommand{\DeepSeekShortName}{DeepSeek}

\newcommand{\GPTLongName}{GPT-4o}
\newcommand{\GPTShortName}{GPT-4o}

\newcommand{\ClaudeLongName}{Claude 3.5 Sonnet}
\newcommand{\ClaudeShortName}{Claude}

\usepackage{times}
\usepackage{latexsym}
\usepackage{enumitem}
\usepackage{graphicx}
\usepackage{subcaption}
\usepackage{pifont} 

\usepackage[T1]{fontenc}

\usepackage[utf8]{inputenc}

\usepackage{microtype}

\usepackage{inconsolata}

\usepackage{graphicx} 
\usepackage{tabularx} 
\usepackage{booktabs} 
\usepackage{multirow}
\usepackage{amsmath}
\usepackage{array}
\usepackage{microtype}
\usepackage{makecell}
\usepackage{xcolor}
\usepackage{listings}
\usepackage{longtable}
\usepackage{adjustbox}
\usepackage{caption}
\usepackage{fancyvrb}
\usepackage[T1]{fontenc}
\usepackage{tikz}
\usepackage{tcolorbox}
\usetikzlibrary{shapes.geometric, arrows}
\definecolor{lightgray}{gray}{0.9}
\definecolor{electricblue}{rgb}{0.49, 0.98, 1}

\usepackage{aaai2026}  
\usepackage{times}  
\usepackage{helvet}  
\usepackage{courier}  
\usepackage[hyphens]{url}  
\usepackage{graphicx} 
\usepackage{natbib}  
\usepackage{caption} 

\usepackage{algorithm}
\usepackage{algorithmic}

\usepackage{newfloat}
\usepackage{listings}
\DeclareCaptionStyle{ruled}{labelfont=normalfont,labelsep=colon,strut=off} 
\floatstyle{ruled}
\newfloat{listing}{tb}{lst}{}
\floatname{listing}{Listing}
\title{Improving Multi-turn Task Completion in Task-Oriented Dialog Systems via Prompt Chaining and Fine-Grained Feedback}
\author{
    Moghis Fereidouni\textsuperscript{\rm 1},
    Md Sajid Ahmed\textsuperscript{\rm 2},
    Adib Mosharrof\textsuperscript{\rm 1},
    A.~B.~Siddique\textsuperscript{\rm 1}
}

\affiliations{
    \textsuperscript{\rm 1}University of Kentucky, 
    \textsuperscript{\rm 2}Independent Researcher\\
    \{moghis.fereidouni, adib.mosharrof, ab.siddique\}@uky.edu, 
    sajid.ahmed1@northsouth.edu
}

\usepackage{bibentry}

\begin{document}

\maketitle


\begin{abstract}
\vspace{-8pt}

Task-oriented dialog~(TOD) systems facilitate users in accomplishing complex, multi-turn tasks through natural language.
While instruction-tuned large language models~(LLMs) have demonstrated strong performance on a range of single-turn NLP tasks, they often struggle with reliable multi-turn task completion in TOD settings, particularly when generating API calls required to interact with external systems. 
To address this, we introduce {\ours}, a novel framework that improves LLM-based TOD systems through \myNum{1}~prompt chaining and \myNum{2}~fine-grained feedback. Prompt chaining enables zero-shot generalization to new domains by automatically synthesizing a schema-aligned in-context example for the target task. 
Fine-grained feedback verifies each generated API call against the domain schema, identifies specific errors, and provides targeted correction prompts. 
To evaluate task completion reliability, we introduce full API Call Accuracy as a robust metric, along with detailed sub-metrics to capture common failure modes. 
We conduct extensive experiments on the SGD and BiTOD benchmarks using four LLMs. 
{\ours} improves Full API accuracy, surpassing state-of-the-art AutoTOD by 37.10\% on SGD and supervised learning-based baseline SimpleTOD by 10.32\% on BiTOD. 
Human evaluations further confirm that LLMs integrated with {\ours} achieve superior task completion, fluency, and informativeness compared to existing methods.

\end{abstract}

\vspace{-20pt}
\section{Introduction}
\vspace{-5pt}

\begin{figure}[t!]
    \centering
\includegraphics[width=0.36\textwidth]{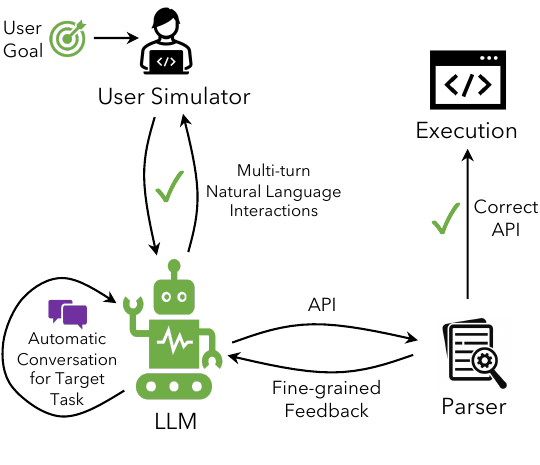} 
    \vspace{-12pt}
\caption{Overview of {\ours}: Prompt chaining circumvents the need for human-curated dialog for each task, and fine-grained feedback loop from the API parser significantly improves task completion rates.}
\vspace{-16pt}
\label{fig:realtod_architecture}
\end{figure}

Task-oriented dialog (TOD) systems enable users to accomplish tasks, such as booking flights, making restaurant reservations, and managing appointments, through multi-turn, natural language interactions \cite{he-etal-2022-space}. 
To successfully complete tasks, these systems must understand user intent, retrieve task-specific information from both user and external systems, and generate coherent responses to guide the user toward task completion (e.g., successful reservation). 
However, traditional TOD systems require extensive domain-specific fine-tuning using manually annotated datasets~\cite{xu2024rethinking, Mi_Wang_Li_2022}, making them laborious and expensive to scale across domains.

Recent advances in large language models~(LLMs)~\cite{Radford2019LanguageMA, brown2020language, Chung2022, shu2024rewritelm} have significantly improved performance across a wide range of single-turn natural language processing~(NLP) tasks, such as text classification \cite{sun-etal-2023-text, wang2023large, zhang2024pushing}, summarization \cite{Pu2023SummarizationI, zhang-etal-2024-benchmarking, van2023clinical}, and machine translation~\cite{cui-etal-2025-multilingual, wang-etal-2023-document-level, gain2025bridging}.
In addition to these single-step tasks, LLMs have also shown strong capabilities in multi-turn, open-domain dialog settings \cite{Thoppilan2022LaMDALM, naveed2023comprehensive, dubey2024llama, achiam2023gpt}, where coherent response generation given dialog context is critical.

Motivated by these advances, recent work has explored leveraging off-the-shelf LLMs for zero-shot TOD settings~\cite{xu2024rethinking}. 
These models can generate fluent and contextually appropriate responses, and with careful prompting, can even be controlled to follow task schemas. 
However, despite their promise, LLMs often struggle with reliable task completion, particularly when generating API calls required to interact with external systems. 
Common issues include hallucinated API results, incorrect method names, missing required parameters, and invalid slot-value pairs \cite{10.5555/3666122.3666499, jain2024mitigating, song2025callnavi}, all of which can lead to execution failures and incomplete tasks. 
These issues reveal a critical gap between LLMs' natural language fluency and their ability to execute tasks reliably in multi-turn dialogs, presenting a major barrier to their practical deployment.

To address these issues, we propose {\ours}, a novel framework that enhances LLM-based TOD systems through \myNum{i}~prompt chaining and \myNum{ii}~fine-grained feedback.
LLMs' performance on single-turn NLP tasks has been shown to improve significantly when provided with in-context examples~\cite{liu-etal-2022-makes, dong-etal-2024-survey, agarwal2024manyshot}.
However, applying this strategy in TOD settings presents a practical challenge: acquiring multi-turn dialogs that are both realistic and schema-compliant for each target domain or task is labor-intensive and does not scale. 
{\ours} addresses this challenge through prompt chaining, which automatically transforms dialogs from a source domain into in-context demonstrations that align with the schema of an unseen target domain.
This approach substantially improves zero-shot generalization to unseen tasks in unseen domains without requiring hand-crafted example dialogs for each task.
In TOD systems, successful task completion often hinges on generating a correct API call that invokes the correct method and includes all required parameters with valid slot-value pairs.
Even small errors in API calls can result in execution failures, ultimately preventing task completion.
To mitigate this, {\ours} incorporates a fine-grained feedback mechanism that verifies each generated API call against the target domain specification, identifies specific errors, and provides targeted correction prompts.
This iterative process significantly improves the reliability of API calls and, in turn, the overall task completion rate.

In addition to improving task completion in TOD systems, we introduce a rigorous evaluation metric for measuring task success. 
Existing metrics focus on n-gram-based similarity metrics~\cite{papineni-etal-2002-bleu,lin-2004-rouge}, which do not reflect whether a task has been successfully executed.
To better align evaluation with practical deployment needs, we adopt full API Call Accuracy as our primary metric. 
An API call is counted as correct only if the method name, parameter names, associated parameter values, and operator match the reference. 
We further decompose this metric into fine-grained sub-metrics, Method Name Accuracy, Parameter Name Accuracy, Parameter Value Accuracy, and Operator Accuracy, to diagnose specific failure modes. 
This evaluation setup captures end-to-end task completion reliability.
On the other hand, we evaluate the quality of natural language responses using BERTScore \cite{zhang2019bertscore}, which assesses semantic similarity between generated and reference responses via contextual embeddings. We also measure diversity \cite{nekvinda-dusek-2021-shades} of the generated text, providing a more robust evaluation than traditional n-gram-based metrics.

We evaluate {\ours} on two benchmark datasets, SGD~\cite{rastogi2020towards} and BiTOD~\cite{lin2021bitod}, using four LLMs: two proprietary models (GPT-4o and Claude) and two open-source models (DeepSeek and LLaMA).
Our results show that {\ours} significantly improves API accuracy across all models and datasets. 
On SGD, it surpasses AutoTOD~\cite{xu2024rethinking} by 37.10\% in full API accuracy, while on BiTOD, it outperforms supervised fine-tuned SimpleTOD~\cite{hosseini2020simple} by 10.32\%. 
Human evaluations confirm that LLMs integrated with {\ours} generate more fluent, informative, and effective task completions than baseline models. 
Our ablation study demonstrates that both prompt chaining and fine-grained feedback contribute to improved multi-turn dialog quality and reliable task completion.

\section{Related Works} 

\stitle{Fine-Tuned Task-Oriented Dialog Systems.} 
TOD systems are typically classified into pipeline-based and end-to-end approaches. Pipeline-based methods \cite{WILLIAMS2007393, lee-2013-structured, LEE2009466, peng-etal-2020-shot, chen-etal-2019-semantically} decompose the system into modular components—natural language understanding, dialog state tracking, policy learning, and natural language generation—allowing independent optimization of each module. In contrast, end-to-end approaches \cite{hosseini2020simple, madotto-etal-2018-mem2seq, su-etal-2022-multi, mosharrof2023zero, SiddiqueTOD, lei-etal-2018-sequicity, lin-etal-2020-mintl, imrattanatrai-fukuda-2023-end} generate responses directly, bypassing these modules. A major drawback of these fine-tuned methods is their reliance on high-quality labeled data, which can be a significant limitation.

\stitle{LLM-Powered Systems.} The rise of LLMs has led to the development of various intelligent systems, which can be broadly categorized into three classes. The first class includes Web Agents, which facilitate online interactions for information retrieval and task execution \cite{yao2023react, kim-etal-2024-prospector, ma2023laser, fereidouni-etal-2024-grounded, NEURIPS2022_82ad13ec, sridhar2023hierarchical, furuta2024multimodal}. The second class consists of Mobile Agents, which focus on optimizing LLM-based decision-making for performing diverse tasks on mobile applications \cite{bai2024digirl, lee2023explore, AutoDroid, wen2023droidbot, wang2024mobile, wang2024mobilev2}. The third and most relevant class to our work is LLM-powered TOD Systems \cite{chung-etal-2023-instructtods, Mi_Wang_Li_2022, gao-etal-2023-adaptive, hudecek-dusek-2023-large, UnravelingChatGPT}. Specifically, AutoTOD \cite{xu2024rethinking} shares similarities with our approach; however, AutoTOD does not account for the possibility of LLMs making errors in generating API calls and lacks proper evaluation of API accuracy.

\stitle{User Simulators.}
One of the earliest data-driven user simulators is \cite{EckertUser}, where user actions are generated probabilistically based on system actions. Additionally, there have been many advancements in data-driven user simulation. For instance,  transformer-based architectures have been leveraged for domain-independent simulation \cite{lin-etal-2021-domain, lin-etal-2022-gentus} with GPT-based models integrating goal state tracking \cite{liu-etal-2022-generative}. Reinforcement learning has also been applied to fine-tune generative simulators \cite{tseng-etal-2021-transferable, cheng-etal-2022-multiwoz}.
More recently, in-context learning with LLMs has enabled user simulation without fine-tuning, \cite{terragni2023context, davidson2023user}. Similar to \cite{lin-etal-2021-domain, lin-etal-2022-gentus, liu-etal-2022-generative}, our user simulator employs transformer architectures.

\vspace{-10pt}
\section{Proposed Framework: {\ours}}
\label{sec:realtod}

We introduce {\ours}, a task-completion-centric, TOD framework that eliminates the need for fine-tuning while seamlessly scaling to new domains.
While recent LLM-based TOD systems can generate fluent and contextually appropriate responses, they often fail to execute tasks reliably, particularly when producing API calls needed to interact with external systems.
To address this, {\ours} introduces two key innovations:
\myNum{1}~prompt chaining, which first transforms a source-domain dialog into a schema-aligned demonstration for a new target task, and then uses the generated dialog as an in-context example, eliminating the need for hand-crafted demonstrations, to enable zero-shot generalization to the target domain on-the-fly; and
\myNum{2}~fine-grained feedback, which verifies each generated API call against the domain schema, identifies specific errors, and provides targeted correction prompts to improve task execution reliability.


\subsection{Problem Formulation} \label{sec:problem_formulation} 

We formulate multi-turn task completion as a conditional sequence generation problem, where the LLM produces natural language responses or API calls to help users achieve their goals across relevant domains. 
Each API call includes the method name, a dictionary of parameter names, and their corresponding values.

Formally, a domain \( d_x \in D \) is characterized by a domain schema, which consists of a set of user intents \( \mathcal{I}_{d_x} \). 
An intent represents a specific goal the user aims to achieve through the multi-turn conversation. 
For example, in the ``Flights'' domain, an intent might be ``Book a Flight''.
Each intent \( i \in \mathcal{I}_{d_x} \) is associated with a set of slots \( \mathcal{S}_i \), where each slot $s$ captures relevant constraints to fulfilling the intent. 
For example, the intent of ``Book a Flight'' may involve slots such as ``departure city'' and ``destination city''. 
We define a slot \( s \) as a tuple:
\[
s = (\texttt{name}(s), \texttt{is\_required}(s), \texttt{values}(s))
\]
where  \( \texttt{name}(.) \) specifies the slot's name (e.g., ``departure city''), \( \texttt{is\_required}(.) \) is a boolean flag indicating whether the slot is mandatory, and \( \texttt{values}(.) \) specifies a predefined set of possible values for categorical slots (e.g., ``business class'', ``economy class'' in the ``Flights'' domain). 
If the slot accepts free-form inputs, this field remains empty.
For brevity, we will refer to the name of a slot \( \texttt{name}(s) \) as \(s_m\).
Formally, the schema for a domain \( d_x \) is represented as:
\[
\Sigma_{d_x} = (d_x, \mathcal{I}_{d_x}, \{ \mathcal{S}_i \mid i \in \mathcal{I}_{d_x} \}).
\]

In addition to generating natural language responses, the model may need to retrieve information from external systems or execute actions via API calls to accurately fulfill a user's goal. Each API call corresponds to a specific intent in a domain and a set of specified constraints, represented as slot-value pairs.
Formally, an API call $a_n$ is defined as:  
\(
a_n = \texttt{API}(\texttt{method} = i,  \texttt{parameters} = \{(s_m , v), (\cdots) \mid s_m \in \mathcal{S}_i\}),
\)
where  \( i \) is the intent,  \( s_m \) is the slot name, and \( v \) is its assigned value.   
For instance, in the ``Flights'' domain, an API call for booking a flight may look like:  
\(
\texttt{API}(\texttt{method} = \text{``Book\_a\_Flight''}, \texttt{parameters} =\{ (\text{``departure\_city''}, \text{``New York''}), (\text{``destination''},\\ \text{``London''}), (\cdots) \}).
\)

A dialog session in a domain \( d_x \) consists of a sequence of user utterances and system responses across multiple turns. 
We define a session \( \mathcal{T}_{d_x} \) of up to \( T \) turns as:  
\[
\mathcal{T}_{d_x} = \bigl((u_1, r_1), (u_2, r_2), \dots, (u_T, r_T)\bigr)
\]
where \( u_t \) is the user's utterance at turn \( t \), \( r_t \) is the system's response at turn \( t \), which can either be a natural language reply or an API call. 
The dialog history up to turn \( t \), denoted as \( H_t \), consists of all previous exchanges up to and including the current user utterance:  
\(
H_t = \{(u_1, r_1), (u_2, r_2), \dots, (u_{t-1}, r_{t-1}), u_t\}.
\)

\subsection{Prompt Chaining} \label{sec:prompt-chain}

In-context examples have been shown to substantially improve LLM performance in a wide range of single-turn NLP tasks~\cite{liu-etal-2022-makes, dong-etal-2024-survey, agarwal2024manyshot}. 
However, their potential in multi-turn, schema-constrained dialog tasks, such as those in TOD systems, remains under-explored. 
Unlike single-step tasks, TOD requires models to track dialog history, collect slot values across multiple turns, and conform to structured API schemas. Manually creating such in-context examples for every target domain is labor-intensive and hinders scalability.
To address this, {\ours} introduces a two-stage prompt chaining mechanism that enables zero-shot generalization to new domains without requiring any hand-crafted dialogs.
Specifically, {\ours} employs a two-stage prompt chaining mechanism, which consists of two sequential prompting phases: \myNum{i}~example dialog generation that transforms an example dialog from a source domain into a target domain while maintaining task-specific consistency; and \myNum{ii}~task adaptation that leverages the generated example dialog for in-context learning in the target domain.

\vspace{5pt}
\stitle{Example Dialog Generation.}
The first phase constructs an example dialog in the target domain by leveraging the schema mapping between the source domain and target domain. 
Formally, the inputs to LLM in this phase include the source domain schema \(\Sigma_{d_x}\), an example dialog \( \mathcal{T}_{d_x} \) in the source domain, an instruction prompt \( P_1 \) specifying the transformation process, and the target domain schema  \(\Sigma_{d_y}\). The output is a new example dialog \( \mathcal{T}_{d_y} \) that aligns with the intents \(\mathcal{I}_{d_x} \) and associated slots \( \mathcal{S}_{i_x} \) in the target domain \( d_y \).

\vspace{5pt}
\stitle{Task Adaptation.}
Once the example dialog \( \mathcal{T}_{d_y} \) in target domain \( d_y \) is generated, the second phase leverages this as an in-context learning example to enhance the model's adaptation in the target domain. 
At each dialog turn \( t \), the inputs to the LLM include the target domain schema \( \Sigma_{d_y} \), the generated example dialog \( \mathcal{T}_{d_y} \), the dialog history up to turn \( t \) (denoted as \( H_t \)), and an instruction prompt \( P_2 \) that guides the response generation process.

The LLM then produces the system response \( r_t \), which can be either a natural language reply or an API call, depending on the current task context.
Since a single dialog may span multiple domains, we can denote the set of target domains involved in a dialog session as \( \{d_1, d_2, \dots, d_m\} \subseteq D \), and extend to the formulation to condition on all relevant domain schemas
\( \{\Sigma_{d_j}\}_{j=1}^{m} \).
  
\vspace{5pt}
\stitle{Instruction Prompts.} 
The prompt \(P_1\) begins with a task description on generating a dialog from a schema, then presents {domain\_X}'s schema and its sample conversation. It instructs the LLM to analyze this structure, apply it to {domain\_Y}, and generate a corresponding conversation. (For the full prompt \(P_1\), see Appendix \ref{sec:prompt1}.) The instruction prompt \(P_2\) consists of two main parts: a task description and general guidelines. It directs the system to collect required slot values before API calls and use search results for accurate responses. The guidelines emphasize limiting slot requests per turn and confirming user inputs before invoking the API call. (For the full prompt \(P_2\), see Appendix \ref{sec:prompt2}.)

\subsection{Fine-Grained Feedback}

In TOD systems, successful task completion often depends on generating precise API calls that invoke the correct method and include all required parameters with well-formed slot-value pairs. 
However, LLMs, despite their fluency, operate in a free-form text generation mode and must reason over a long dialog history, while adhering to rigid schema constraints. 
This often results in API calls that are incomplete, inconsistent, or even invalid, issues that cannot be reliably prevented through prompting alone.
To minimize these errors and ensure successful task execution, {\ours} integrates a fine-grained feedback mechanism via a generic API parser. Given a domain schema \( \Sigma_{d_x} \) and an API call \(a\), the parser verifies the correctness of the request before execution. 
If the API call conforms to the schema, it is passed for execution; otherwise, the parser provides fine-grained feedback to the LLM for correction.  
The verification process identifies three types of errors: \myNum{i}~incorrect method name, where the API method does not match any intent \( i \not\in \mathcal{I}_d \); \myNum{ii}~incorrect slot name, where a provided slot is not defined in the schema \( s_m \not\in \mathcal{S}_i \) for the given intent; and \myNum{iii}~missing required slots, where required slots \( s_m \) with \( \texttt{is\_required}(s_m) = \text{True} \) are absent in the API parameters. 
Upon detecting an error, the parser returns fine-grained feedback specifying the issue, allowing the LLM to correct its response.

\vspace{-8pt}
\section{User Simulator}

Evaluating a TOD system ideally requires interactions with real users to assess its effectiveness in goal-oriented scenarios. 
However, deploying and managing real-user evaluations is expensive and time-consuming. 
To overcome this challenge, we develop a goal-driven user simulator that can interact with TOD systems in a controlled and scalable manner.
An effective user simulator must first accurately convey its needs by specifying the required slot values (e.g., `departure city'') before optionally requesting information (e.g., the flight's arrival time'') from the TOD system. 
To construct such a simulator, we utilize dialog data \( \mathcal{T}_{d_x} \)  consisting of user goals, expressed through API calls \( A = [a_1, a_2, \dots, a_n] \), and the request slots  \( R =  [s_1, s_2, \dots, s_m] \) that the user should request. 
To train the user simulator, we optimize an instruction-finetuned model as:
\vspace{-4pt}
\[
\mathcal{L} = - \sum_{k=1}^{|u_t|} \log p(w_k \mid w_{<k}, H_t, A, R),
\]
where \( w_k \) denotes the \( k \)-th token in the user utterance \( u_t \) at turn \(t\), and \( w_{<k} \) represents all preceding tokens in the same utterance. 
The simulator learns to express user goals in natural language by answering requested slots from the TOD system, and requests additional information from the TOD system, conditioned on the set of API calls \(A\), request slots \(R\), and dialog context \(H_t\).
To conduct an interactive session between a trained user simulator and the TOD system, the simulator initiates the conversation by retrieving the first user goal \( a_1 \) from \(A\) and associated request slot \( s_1 \) from \(R\).
This process continues iteratively until all user goals in 
A and their associated request slots in R have been processed.

\section{Experiments}




\subsection{Datasets}
We conduct our experiments using two datasets: the Schema-Guided dialog (SGD) dataset \cite{rastogi2020towards} and the Bilingual Task-Oriented dialog (BiToD) dataset \cite{lin2021bitod}. Since BiToD includes dialogs in both Chinese and English, we retain only the English dialogs for our analysis. Both datasets provide domain-specific schemas along with corresponding dialog conversations, which are essential for baseline models. A comparative summary of key statistics for both datasets is presented in Table \ref{tab:sgd_bitod_comparison}.


\begin{table}[t]
    \centering
    \renewcommand{\arraystretch}{1.2}
    \begin{adjustbox}{max width=0.40\textwidth}
        \begin{tabular}{lcc}
            \toprule
            \textbf{Statistic} & \textbf{SGD} & \textbf{Bitod} \\
            \midrule
            Total Dialogs & 4,201 & 352 \\
            Total Dialogs (Single-domain) & 1,331 & 111 \\
            Total Dialogs (Multi-domain) & 2,870 & 241 \\ 
            Total API Calls & 13,239 & 1,005 \\
            Total API Calls (Single-domain) & 2,188 & 127 \\
            Total API Calls (Multi-domain) & 11,051 & 878 \\
            Total Turns & 89,428 & 6,979 \\ 
            Total User Req. Slots & 8,271 & 500 \\ \hline
            Avg. API calls per dialog & 3.15 & 2.85 \\
            Avg. API calls (Single-domain) & 1.64 & 1.14 \\
            Avg. API calls (Multi-domain) & 3.85 & 3.64 \\
            Avg. turns per dialog & 21.28 & 19.82 \\
            Avg. User Req. Slots & 1.96 & 1.42 \\
            Avg. parameters per API call & 2.96 & 3.51\\ \hline
            Total Unique API methods & 34 & 7\\ 
            Total Unique API parameters & 88 & 20\\ 
            \bottomrule
        \end{tabular}
    \end{adjustbox}
    \caption{Test Dataset Statistics for SGD and BiTOD.
    }
    \label{tab:sgd_bitod_comparison}
    \vspace{-10pt}
\end{table}

\subsection{Experimental Setup} 

We integrated four LLMs in {\ours}: two open-source models, \DeepSeekLongName\ \cite{liu2024deepseek} and \LlamaLongName\ \cite{dubey2024llama}, and two proprietary models, \GPTLongName\ \cite{achiam2023gpt} and \ClaudeLongName\ \cite{anthropic2023claude}. For \GPTLongName, we accessed the model via the official OpenAI API, while \ClaudeLongName\ was queried using the official Anthropic API.

We fine-tune Flan-T5 model~\cite{Chung2022} to act as a user simulator for each dataset. Specifically, we use the "google/flan-t5-base" model, consisting of 250 million parameters.
During fine-tuning, we set the warm-up steps to 100 and applied early stopping with patience of three. The models were trained for 10 epochs.


\subsection{Evaluation Metrics}
To comprehensively evaluate the performance of {\ours} and baseline models, we assess the following: \myNum{i}~Dialog-Level System Response, \myNum{ii}~Dialog-Level Language Diversity, \myNum{iii}~API Call, and \myNum{iv}~Dialog Success Rate.


\stitle{Dialog-Level System Response.}
To assess the quality of the responses generated by {\ours}, we removed all user responses produced by our user simulator, retaining only system responses. We then concatenated all system turns into a single text containing only system-generated outputs. The same process was applied to the ground truth dialog, keeping and concatenating only the system turns. Finally, we evaluated system response quality at the dialog level by comparing the generated responses to the ground truth using BERTScore~\cite{zhang2019bertscore}, a metric that measures semantic similarity between texts. Furthermore, we utilize "microsoft/mpnet-base" as the foundational model for computing BERTScore.


\stitle{Dialog-Level Language Diversity.}
To assess the lexical diversity of the system responses, we compute Shannon Entropy (SE) and Bigram Conditional Entropy  (CE) over the system turns \cite{terragni2023context,xu2024rethinking}. Following the same preprocessing steps as in the previous section, we remove all user turns and concatenate only the system responses for each dialog. The SE and CE are then computed at the dialog level to quantify the diversity and richness of the language used by the model.

\stitle{API Calls.}
To evaluate the quality of API Calls, we first extract the key-value pairs $ (name(s_k),v_k)_{k=1}^n $, along with method name $i$ from the generated API call using regular expressions. \emph{Method Accuracy} evaluates whether the generated API call uses the correct method name, assessed using exact matching. \emph{Parameter Name Accuracy} determines whether all ground truth key names are included in the generated API call, using fuzzy matching. \emph{Parameter Value Accuracy} verifies if the value associated with a correctly predicted key matches the ground truth, also using fuzzy matching. Notably, this metric is computed only when the corresponding \emph{Parameter Name} is correctly predicted. \emph{Operator Accuracy} applies specifically to the BiToD dataset, as only this dataset includes API calls with operators (e.g., ``at\_least'', ``one\_of''). We assess this using fuzzy matching. \emph{Full API Accuracy} measures whether the entire API call -- including the method, parameter, values, and, for BiToD, the operator -- matches the ground truth.





\stitle{Dialog Success Rate.} This metric measures the percentage of dialogs in which all API calls achieve 100\% \emph{Full API Accuracy}. In other words, it represents the proportion of dialogs where every generated API call matches the ground truth, ensuring complete correctness throughout the dialog.

\subsection{Baseline Methods}


We compare {\ours} against several strong baselines. 

\textbf{SyncTOD} \cite{saley-etal-2024-synergizing} improves in-context learning for LLM-powered task-oriented dialog systems by using an auxiliary model to predict hints about the expected response, which guide the selection of proper in-context exemplars.


\textbf{AutoTOD} \cite{xu2024rethinking} is a zero-shot task-oriented dialog agent that eliminates traditional modules, relying only on instruction-following LLMs like GPT-4. It requires no task-specific training and autonomously decides actions, queries APIs, and generates responses.


\textbf{ZS-TOD} \cite{mosharrof2023zero} is a zero-shot task-oriented dialog system that generalizes to unseen domains using domain schemas instead of memorizing task-specific patterns. It replaces full dialog history with a concise summary, reducing context complexity.

\textbf{Q-TOD} \cite{tian-etal-2022-q} is a query-driven task-oriented dialog system that employs a Transformer to generate natural language queries from the dialog context for retrieving relevant knowledge, which is then used to generate system responses.



\textbf{SOLOIST} \cite{peng2021soloist} is a Transformer-based task-oriented dialog system that unifies multiple dialog modules into a single pre-trained model. It leverages transfer learning and machine teaching, allowing adaptation to new tasks with minimal labeled data.


\textbf{SimpleTOD} \cite{hosseini2020simple} treats task-oriented dialog as a single sequence generation problem, using a causal language model to predict dialog state, actions, and responses auto regressively.


\begin{table*}[t!]
\centering
\resizebox{\textwidth}{!}{ 
\begin{tabular}{l|l|ccc|ccc|ccc|ccc|ccc}
\toprule

\textbf{Dataset} & \textbf{LLM Model} & 
\multicolumn{3}{c|}{\makecell{\textbf{Method} \\ \textbf{Accuracy}}} & 
\multicolumn{3}{c|}{\makecell{\textbf{Param Names} \\ \textbf{Accuracy}}} & 
\multicolumn{3}{c|}{\makecell{\textbf{Param Values} \\ \textbf{Accuracy}}} & 
\multicolumn{3}{c|}{\makecell{\textbf{Operator} \\ \textbf{Accuracy}}} & 
\multicolumn{3}{c}{\makecell{\textbf{Full API} \\ \textbf{Accuracy}}} \\ 
\cmidrule(lr){3-5} \cmidrule(lr){6-8} \cmidrule(lr){9-11} \cmidrule(lr){12-14} \cmidrule(lr){15-17}

& & \makecell{\textbf{Single}} & \makecell{\textbf{Multi}} & \makecell{\textbf{Both}} & 
\makecell{\textbf{Single}} & \makecell{\textbf{Multi}} & \makecell{\textbf{Both}} & 
\makecell{\textbf{Single}} & \makecell{\textbf{Multi}} & \makecell{\textbf{Both}} & 
\makecell{\textbf{Single}} & \makecell{\textbf{Multi}} & \makecell{\textbf{Both}} & 
\makecell{\textbf{Single}} & \makecell{\textbf{Multi}} & \makecell{\textbf{Both}} \\ 
\midrule

\multirow{8}{*}{\textbf{SGD}}
 & SOLOIST                & 61.56 & 64.98 & 64.41 & 44.88 & 47.45 & 47.02 & 42.99 & 45.50 & 45.08 &  &  &  & 24.54 & 26.58 & 26.24 \\ 
 & Q-TOD                & 40.08 & 42.64 & 42.21 & 31.91 & 35.03 & 34.51 & 31.35 & 34.39 & 33.88 &  &  &  & 19.24 & 22.18 & 21.69 \\
 & SimpleTOD              & 53.51 & 59.65 & 58.62 & 44.32 &  50.16 & 49.18 & 41.92 & 47.38 & 46.46 & N/A & N/A & N/A & 17.55 & 20.70 & 21.25 \\ 
 & ZS-TOD                 & 29.11 & 50.46 & 46.88 & 21.52 & 38.26 & 35.46 & 21.52 & 34.87 & 32.27 &  &  &  & 07.67 & 14.70 & 13.52 \\ 
 & SyncTOD                & 30.80 & 46.93 & 44.232 & 27.60 & 46.26 & 43.13 & 26.53 & 44.57 & 41.55 &  &  &  & 16.36 & 30.92 & 28.48 \\
 & AutoTOD                & 56.67 & 62.61 & 61.61 & 58.50 & 65.05 & 63.95 & 54.74 & 61.51 & 60.37 &  &  &  & 42.04 & 48.16 & 47.14 \\ 
 \cmidrule(){2-17}
 & {\ours}-{\small \GPTShortName}                 & 80.60 & \underline{71.26} & \underline{72.81} & 84.72 & \underline{73.54} & \underline{75.40} & 81.44 & \underline{70.57} & \underline{72.38} &  &  &  & \underline{68.71} & \underline{57.89} & \underline{59.69} \\ 
 & {\ours}-{\small \ClaudeShortName}             & \textbf{88.29} & \textbf{81.74} & \textbf{82.83} & \textbf{88.91} & \textbf{80.59} & \textbf{81.97} & \textbf{85.66} & \textbf{77.08} & \textbf{78.51} & N/A & N/A & N/A & \textbf{72.58} & \textbf{63.04} & \textbf{64.63} \\ 
 &  {\ours}-{\small \DeepSeekShortName}            & \underline{81.11} & 69.25 & 71.22 & 83.78 & 70.80 & 72.96 & 79.42 & 66.42 & 68.58 &  &  &  & 63.27 & 51.67 & 53.59 \\ 
 & {\ours}-{\small \LlamaShortName}              & 80.32 & 69.60 & 71.39 & \underline{86.65} & 71.46 & 73.99 & \underline{82.36} & 67.56 & 70.02 &  &  &  & 63.23 & 52.58 & 54.35 \\ 
\midrule

\multirow{8}{*}{\textbf{BiTOD}}
 & SOLOIST                & 52.5 & 69.82 & 67.97 & 27.96 & 61.37 & 57.81 & 27.18 & 59.02 & 55.63 & 27.96 & 61.11 & 57.58 & 20.00 & \underline{50.81} & 47.53 \\ 
 & Q-TOD                & 44.09 & 42.74 & 42.89 & 44.56 & 43.42 & 43.54 & 44.54 & 42.58 & 42.79 & 44.56 & 43.37 & 43.50 & 42.23 & 39.14 & 39.47 \\
 & SimpleTOD              & 28.57 & 59.71 & 56.39 & 27.61 & 59.19 & 55.82 & 26.83 & 57.32 & 54.07 & 27.61 & 59.01 & 55.66 & 20.49 & \textbf{50.89} & 47.64 \\ 
 & ZS-TOD                 & 26.0 & 33.53 & 32.73 & 23.11 & 31.55 & 30.65 & 23.11 & 30.23 & 29.47 & 22.49 & 29.86 & 29.07 & 16.14 & 20.32 & 19.88 \\ 
 & SyncTOD                & 36.64 &  31.70 & 32.23 & 60.80 & 41.79 & 43.84 & 55.68 & 38.86 & 40.67 & 52.42 & 37.12 & 38.77 & 13.66 & 17.35 & 16.95 \\
 & AutoTOD                & 64.29 & 48.42 & 49.84 & 41.73 & 23.84 & 25.44 & 38.45 & 21.84 & 23.33 & 31.48 & 21.15 & 22.08 & 17.86 & 14.04 & 14.38 \\
  \cmidrule(){2-17}
 & {\ours}-{\small \GPTShortName}                & 82.81 & \underline{73.82} & 74.97 & 82.81 & \underline{71.20} & 72.69 & 82.31 & \textbf{66.90} & \underline{68.88} & 79.56 & 70.55 & 71.71 & \underline{68.75} & 50.17 & \textbf{52.56} \\ 
 & {\ours}-{\small \ClaudeShortName}            & \textbf{94.49} & 66.74 & \textbf{83.07} & \textbf{91.47} & 69.06 & \textbf{79.15} & \textbf{90.05} & 60.19 & \textbf{73.34} & \textbf{90.81} & \textbf{76.38} & \textbf{78.22} & \textbf{71.65} & 47.30 & \underline{50.40} \\ 
 & {\ours}-{\small \DeepSeekShortName}            & \underline{90.55} & \textbf{75.92} & \underline{77.79} & \underline{86.93} & \textbf{72.02} & \underline{73.92} & \underline{84.69} & \underline{65.12} & 67.62 & \underline{85.33} & \underline{71.35} & \underline{73.13} & 62.20 & 44.47 & 46.73 \\ 
 & {\ours}-{\small \LlamaShortName}             & 85.16 & 61.78 & 64.81 & 83.46 & 60.84 & 63.77 & 82.43 & 54.16 & 57.82 & 79.27 & 59.19 & 61.78 & 66.41 & 35.88 & 39.83 \\ 
\bottomrule

\end{tabular}%
}
\caption{API Call Accuracy breakdown across all models on the SGD and BiTOD datasets. Accuracy is reported across multiple metrics, including method, parameter name, parameter value, operator, and overall full API accuracy. Results are shown for single-domain, multi-domain, and both domains.}
\label{table_api}
\vspace{-10pt}
\end{table*}

\vspace{-8pt}
\section{Results and Analysis}

Here, we focus on the accuracy of API calls generated by the dialog system and the overall quality of system responses at the dialog level. A fine-grained evaluation metric for system responses is provided in Appendix~\ref{sec:inform_accuracy_metric}. For details on user simulator performance, see Appendix~\ref{sec:User_Simulator_Performance}.

\subsection{Evaluating the Quality of API Calls}
Table~\ref{table_api} presents the API call accuracy results on both the SGD and BiToD datasets. 

\stitle{Comparing \ours{} Performance to Baselines.} 
A key observation is that across both datasets, nearly all variants of {\ours} outperform the baseline models across all evaluation metrics, including Method Accuracy, Param Names Accuracy, Param Values Accuracy, Operator Accuracy (for BiToD), and Full API Accuracy. Notably, when focusing on Full API Accuracy, we see substantial gains of {\ours} over baselines. For instance, \ClaudeShortName\ surpasses AutoTOD, the strongest baseline, by 37.10\% in Full API Accuracy on the SGD dataset. Similarly, on BiToD, \GPTShortName\ outperforms SimpleTOD, the best baseline model, by 10.32\%, highlighting the robustness of our approach. Moreover, to view the dialogs generated by {\ours}, please refer to Appendix~\ref{sec:Example_Dialog_Responses}.

\stitle{Open-Source vs.\ Proprietary Models.} A notable trend in Table~\ref{table_api} is the consistent superiority of proprietary models (\GPTShortName, \ClaudeShortName) over open-source counterparts (\DeepSeekShortName, \LlamaShortName) in terms of Full API Accuracy across both datasets. For instance, on the SGD dataset, \ClaudeShortName\ outperforms \DeepSeekShortName\ by 20.60\%, while on BiTOD, \GPTShortName\ achieves a 31.96\% higher Full API Accuracy than \LlamaShortName. These results underscore the performance gap between proprietary and open-source LLMs.

\stitle{Model-Specific Observations.} Interestingly, when comparing \LlamaShortName\ and \DeepSeekShortName\ in the Table~\ref{table_api}, their relative performance depends on the dataset. While \LlamaShortName\ yields higher accuracies in most metrics on SGD, the trend reverses in BiToD, where \DeepSeekShortName\ significantly outperforms \LlamaShortName\ on almost all metrics. We attribute this to \DeepSeekShortName’s closer alignment with Chinese data, which proves advantageous for BiToD’s English subset that still contains Chinese references (e.g., restaurant names). This shows that LLM performance in TOD tasks depends on alignment with the dataset's language and domain.

\stitle{Comparison Between Metrics.} Across both datasets, we observe that all models, including baselines and {\ours}, tend to perform better on Method Accuracy and Parameter Names Accuracy than on Parameter Values Accuracy and Operator Accuracy. This suggests that identifying the correct method or parameter name from the domain schema is generally easier than generating the appropriate value or selecting the correct operator.




\subsection{Evaluating the Quality of System Responses} 
So far, we have evaluated the dialog systems primarily based on their ability to generate accurate API calls. However, assessing the overall performance of the dialog systems also requires analyzing the quality of their natural language responses. To this end, we refer to Table~\ref{table_score_inform}. This table shows that SyncTOD consistently outperforms other methods in terms of BERTScore (F1) across both the SGD and BiTOD datasets. This is expected, as SyncTOD is designed to produce responses that closely align with the style of the source dataset. Following SyncTOD, the supervised models rank second in BERTScore (F1), which is again unsurprising given that they are trained to mimic the language patterns found in the training data.

However, a high BERTScore does not necessarily indicate that a model generates high-quality responses. It primarily reflects surface-level similarity to reference responses. Therefore, to evaluate the richness of the generated text, we also consider diversity metrics. As shown in Table~\ref{table_score_inform}, {\ours} clearly excels in this regard, achieving the highest scores in both Shannon Entropy (SE) and Bigram Conditional Entropy (CE). This suggests that while {\ours}’s responses may be slightly less similar to the reference responses (as reflected by its lower BERTScores), they are more varied and diverse. Such diversity is a valuable trait in dialog systems, as it helps prevent repetitive or overly generic replies.

\begin{table}[t!]
\centering

\resizebox{1.0\linewidth}{!}{ 
\begin{tabular}{l|l|ccc|cc}
\toprule

\textbf{Dataset} & \textbf{LLM Model} & 
\multicolumn{3}{c|}{\makecell{\textbf{BERTScore} \\ \textbf{System (F1)}}} &   \multicolumn{2}{c}{\makecell{\textbf{Language Diversity}}}  \\ 
\cmidrule(lr){3-5} \cmidrule(lr){6-7}

& & 
\makecell{\textbf{Single}} & \makecell{\textbf{Multi}} & \makecell{\textbf{Both}} & \makecell{\textbf{SE}} & \makecell{\textbf{CE}} \\ 
\midrule

\multirow{8}{*}{\textbf{SGD}}
 & SOLOIST  & 0.6732   & 0.6811   & 0.6732  & 6.2337  & 0.8688  \\ 
 & Q-TOD  & 0.6559  & \underline{0.7281}   & \underline{0.7049}  & 4.7595 & 0.5543  \\
 & SimpleTOD  & 0.5963  & 0.6227   & 0.6142 & 5.7914 & 0.7282  \\ 
 & ZS-TOD  & 0.5709  & 0.5748  & 0.5736  & 5.5656 & 0.8516   \\ 
 & SyncTOD  & \textbf{0.6867}  & \textbf{0.7566}   & \textbf{0.7341}  & 5.5451 & 0.7403 \\
 & AutoTOD   & \underline{0.6735}   & 0.7080  & 0.6969  & \underline{6.4834} & 1.1753  \\  
 \cmidrule(){2-7}
 & {\ours}-{\small \GPTShortName}  & 0.6547 & 0.6544 &  0.6545 & 6.3101 & 1.0610 \\ 
 & {\ours}-{\small \ClaudeShortName}   & 0.6552 & 0.6694 & 0.6649  & \textbf{6.7078} & \underline{1.3033} \\ 
 & {\ours}-{\small \DeepSeekShortName}   & 0.6345 &  0.6384   &  0.6372 & 6.5418 & 1.0674 \\ 
 & {\ours}-{\small \LlamaShortName}    & 0.6019 & 0.5979 & 0.5992 & 6.4796 & \textbf{1.4515}  \\ 
\midrule

\multirow{8}{*}{\textbf{BiTOD}} 
 & SOLOIST   & 0.5692  & \underline{0.7876}   & 0.7186  & 5.7138 & 0.8909    \\ 
 & Q-TOD  & 0.6481  & 0.7356  & 0.7077 & 4.8532 & 0.6192  \\
 & SimpleTOD     & 0.5781 & \textbf{0.7964}  & \underline{0.7271}  & 5.4943 & 0.8033  \\ 
 & ZS-TOD   & \underline{0.6563}  & 0.7467   & 0.7180  & 5.7151 & 0.7669 \\ 
 & SyncTOD  & \textbf{0.6936}   & 0.7552   & \textbf{0.7355}    & 6.4828 & 1.0154 \\
 & AutoTOD    & 0.6526 & 0.6745   & 0.6675  &  6.5369 & 1.1007  \\  
  \cmidrule(){2-7}
 & {\ours}-{\small \GPTShortName} & 0.6543 &  0.6447 & 0.6477  & 6.4121 & 1.0897 \\ 
 & {\ours}-{\small \ClaudeShortName} & 0.6523 & 0.6392 &  0.6434 & \textbf{6.8699} & \underline{1.3602} \\ 
 & {\ours}-{\small \DeepSeekShortName}   & 0.6454 & 0.6278 & 0.6334 & 6.5743 & 1.0689 \\ 
 & {\ours}-{\small \LlamaShortName}   & 0.5927 & 0.5694 & 0.5769 & \underline{6.5957} & \textbf{1.4415} \\ 
\bottomrule

\end{tabular}%
}
\caption{Comparison of {\ours} with baseline models for Language Diversity and BERTScore for system response on SGD and BiTOD datasets.}
\vspace{-12pt}
\label{table_score_inform}
\end{table}

\subsection{Ablation Study}
To assess the effectiveness of our proposed components, Fine-Grained Feedback and Prompt Chaining, we conducted an ablation study using 100 dialog conversations sampled from the SGD dataset (50 from multi-domain and 50 from single-domain). We evaluated all four variants of {\ours} (\GPTShortName, \ClaudeShortName, \LlamaShortName, and \DeepSeekShortName) under four different settings: one without either of the components, one with Fine-Grained Feedback only, one with Prompt Chaining only, and one with both components. Moreover, we used the Full API Accuracy as our comparison metric. This experimental design allowed us to isolate the impact of each component and determine their individual and combined contributions to performance. The results are provided in the Table \ref{tab:ablation_study}. As it can be seen in Table~\ref{tab:ablation_study}, adding Fine-Grained Feedback alone leads to moderate improvements, indicating its role in refining APIs. Prompt Chaining, on the other hand, provides a more substantial boost. The combination of both components yields the highest accuracy, demonstrating their complementary nature.

\begin{table}[t!]
    \centering
    \resizebox{1.0\linewidth}{!}{ 
    \begin{tabular}{cc|cccc}
        \toprule
        \makecell{\textbf{Fine-Grained} \\ \textbf{Feedback}} & \makecell{\textbf{Prompt} \\ \textbf{Chaining}} & \textbf{\GPTShortName} & \textbf{\ClaudeShortName} & \textbf{\DeepSeekShortName} & \textbf{\LlamaShortName} \\
        \midrule
        \xmark & \xmark & 51.89 & 29.92 & 45.83 & 49.62 \\
        \cmark & \xmark & 56.06 & 36.74 & 48.10& 57.19 \\
        \xmark & \cmark & \underline{64.01} & \underline{70.45} &  \underline{54.92} & \underline{59.84} \\
        \cmark & \cmark & \textbf{66.66} & \textbf{72.34} &  \textbf{59.46} & \textbf{63.63} \\

        \bottomrule
    \end{tabular}
    }
    \caption{Ablation study of full API accuracy on the SGD dataset to evaluate the impact of fine-grained feedback and prompt chaining across different LLMs.}
    \label{tab:ablation_study}
\end{table}

\begin{figure}[t!]
   \centering
   \includegraphics[width=1.0\linewidth]{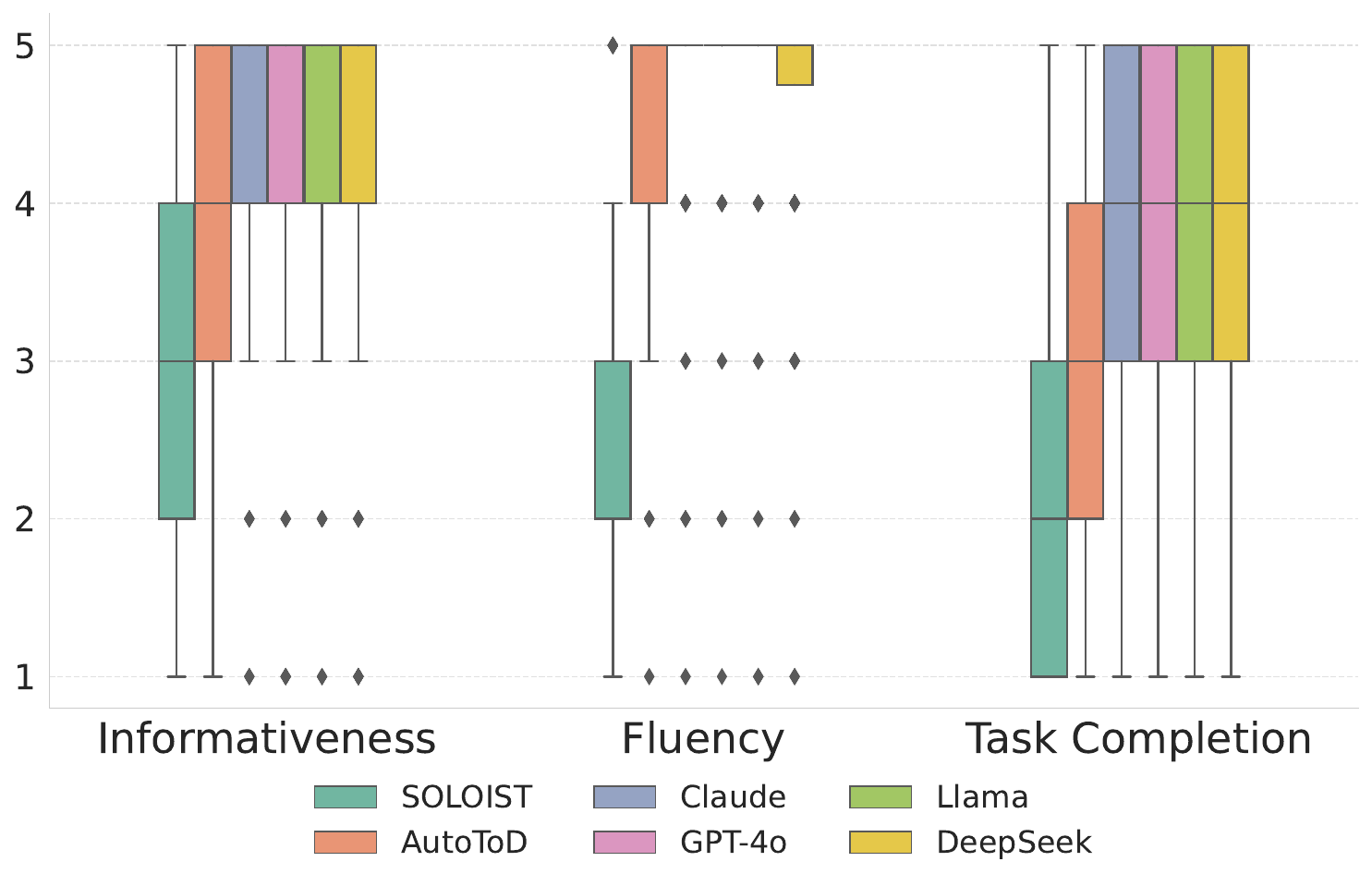}
   \vspace{-16pt}
   \caption{
       Human Evaluation Results on SGD and BiTOD. Evaluators rated generated conversations from 1 to 5 based on three aspects: informativeness, fluency, and task completion.
   }
   \label{fig:human_evaluation}
   \vspace{-6pt}
\end{figure}

\subsection{Human Evaluation}

We conducted a human evaluation using Amazon Mechanical Turk to assess the performance of our models. As baselines, we selected SOLOIST and AutoTOD, based on their good performance in Table~\ref{table_api}, and compared them against all four variants of {\ours}.  For our evaluation, we sampled 100 dialogs from the test sets of our chosen datasets (SGD and BiToD), with 50 each from single and multi-domain tasks. We asked the human evaluators to rate the generated conversations on a scale of 1 to 5 across three key aspects: \emph{Informativeness}, \emph{Fluency}, and \emph{Task Completion Rate}. Figure \ref{fig:human_evaluation} shows the human evaluation results, where all four variants of {\ours} outperformed the baseline models (SOLOIST and AutoToD), supporting the reliability of our evaluation metrics. 

\begin{figure}[t!]
    \centering
    \includegraphics[width=0.5\textwidth]{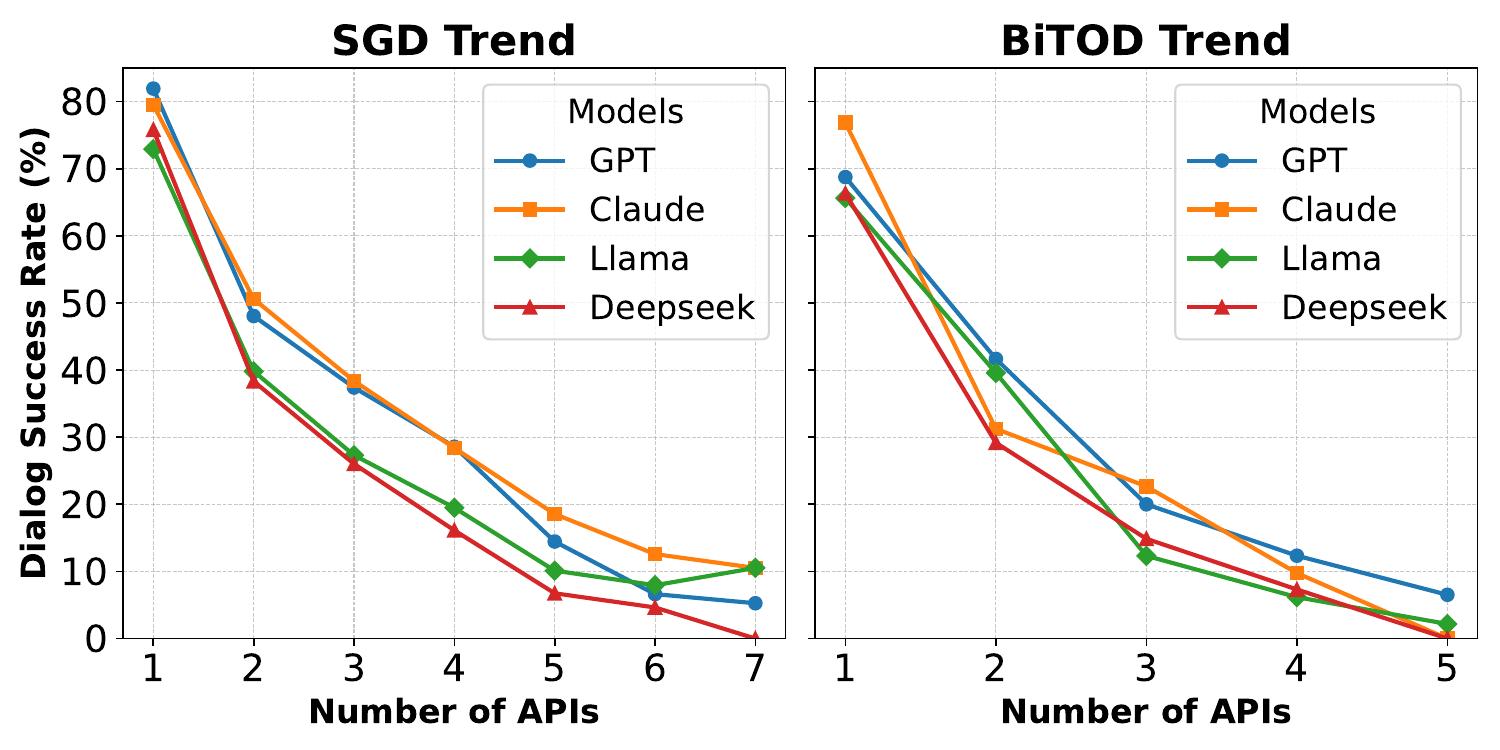}
    \caption{Dialog success rate declines as the number of API calls increases across models on the SGD and BiTOD datasets. This trend underscores the challenge of error propagation in LLM-powered TOD systems, where early mistakes adversely affect later interactions.}
    \label{fig:trend}
    \vspace{-10pt}
\end{figure}

\subsection{Dialog Success Rate} 
To rigorously assess the quality of the generated dialogs, we conducted an experiment to measure the dialog success rate as the number of API calls within a dialog increases. As shown in Figure \ref{fig:trend}, all variants of {\ours} exhibit a declining trend in dialog success rate as the number of API calls increases. This trend is consistent across both the SGD and BiTOD datasets. The primary reason for this decline is the interdependence of API calls. For example, when a user books a restaurant at a particular destination, the same location is often referenced for booking a taxi or later searching for nearby hotels. Any errors in earlier API calls can propagate, making subsequent calls more prone to failure. The Figure \ref{fig:trend} highlights a limitation of LLM-powered TOD systems, suggesting that they are still far from achieving perfect performance. Further research is needed to enhance their ability to handle these scenarios more effectively.

\vspace{-5pt}
\section{Conclusion} 
\vspace{-5pt}

We presented {\ours}, a task-completion-centric framework for TOD systems that eliminates the need for domain-specific fine-tuning and enables reliable performance across diverse domains. 
Our work tackles a fundamental limitation of current LLM-based TOD systems: their inability to consistently complete tasks due to unreliable API call generation, despite their fluency in natural language responses. 
{\ours} introduces three key innovations: \myNum{1}~a prompt chaining mechanism that enables zero-shot generalization to new tasks and domains by automatically generating schema-aligned in-context demonstrations; \myNum{2}~a fine-grained feedback loop that verifies each API call against the domain schema and provides targeted prompts to guide correction, significantly improving task execution reliability; and \myNum{3}~a task-centric evaluation metric that goes beyond traditional n-gram-based metrics by precisely assessing task completion through full API call accuracy and detailed error breakdowns.
Through comprehensive experiments across two benchmark datasets and four LLMs, we demonstrate that {\ours} achieves substantial gains in task success rates, setting a new standard for LLM-based TOD systems. 
Beyond these performance improvements, our analysis reveals persistent challenges in dialog-based task completion, including domain coverage errors in multi-domain settings, failures in long-horizon planning across turns, and misalignment when integrating external search results, highlighting directions for future research.

\bibliography{aaai2026}


\appendix

\setcounter{secnumdepth}{1}
\appendix
\clearpage
\section{Inform Accuracy}
\label{sec:inform_accuracy_metric}


\stitle{Inform Accuracy Metric.}
To evaluate how effectively {\ours} informs the user about the requested slots, we implemented a regex-based system. First, we identify the slots requested by the user and extract their corresponding values from the search results. Then, we use regex matching to determine whether the system’s subsequent responses include those extracted values. If a system turn contains the requested slot values, we consider the system to have successfully provided the required information.

\stitle{Analysis on Inform Accuracy Results.}
As expected, Table~\ref{table_inform_accuracy} shows that LLM-powered models perform strongly on the Inform Accuracy metric. On the SGD dataset, our method {\ours}-{\small \GPTShortName} achieves the highest Inform Accuracy across the Single, Multi, and Both domain settings, with {\ours}-{\small \DeepSeekShortName} ranking second. On the BiToD dataset, however, AutoTOD outperforms all other models. These results reinforce that LLMs, especially when supported by a dedicated system architecture such as {\ours}, are highly effective at accurately providing requested slot values.


\begin{table}[t!]
\centering
\resizebox{1.0\linewidth}{!}{
\begin{tabular}{l|l|ccc}
\toprule
\textbf{Dataset} & \textbf{LLM Model} & \multicolumn{3}{c}{\textbf{Inform Accuracy (\%)}} \\
\cmidrule(lr){3-5}
& & \textbf{Single} & \textbf{Multi} & \textbf{Both} \\
\midrule
\multirow{11}{*}{\textbf{SGD}}
 & SOLOIST                            & 44.54 & 54.42 & 52.66 \\
 & Q-TOD                              & 25.28 & 26.75 & 26.49 \\
 & SimpleTOD                         & 32.15 & 51.11 & 47.74 \\
 & ZS-TOD                             & 12.32 & 11.20 & 11.40 \\
 & SyncTOD                            & 75.00 & 73.12 & 73.45 \\
 & AutoTOD                           & 76.34 & \underline{75.84} & 75.93 \\
 \cmidrule(){2-5}
 & {\ours}-{\small \GPTShortName}     & \textbf{82.93} & \textbf{76.89} & \textbf{78.03} \\
 & {\ours}-{\small \ClaudeShortName}  & 79.44 & 71.37 & 72.85 \\
 & {\ours}-{\small \DeepSeekShortName}& \underline{82.81} & 75.45 & \underline{76.88} \\
 & {\ours}-{\small \LlamaShortName}   & 73.49 & 70.26 & 70.88 \\
\midrule
\multirow{11}{*}{\textbf{BiTOD}}
 & SOLOIST                            & 80.00 & 69.42 & 70.62 \\
 & Q-TOD                              & 35.00 & 26.11 & 27.11 \\
 & SimpleTOD                         & 85.00 & 71.33 & 72.88 \\
 & ZS-TOD                             & 60.00 & 66.24 & 65.53 \\
 & SyncTOD                            & \underline{87.50} & \underline{80.25} & \underline{81.07} \\
 & AutoTOD                           & \textbf{97.50} & \textbf{84.39} & \textbf{85.87} \\
 \cmidrule(){2-5}
 & {\ours}-{\small \GPTShortName}     & 85.18 & 79.78 & 80.26 \\
 & {\ours}-{\small \ClaudeShortName}  & 64.70 & 63.51 & 63.61 \\
 & {\ours}-{\small \DeepSeekShortName}& 86.66 & 79.25 & 79.93 \\
 & {\ours}-{\small \LlamaShortName}   & 64.28 & 65.06 & 65.00 \\
\bottomrule
\end{tabular}}
\caption{Comparison of Inform Accuracy (\%) on SGD and BiTOD datasets across all baseline models and {\ours}.}
\label{table_inform_accuracy}
\end{table}

\section{Limitations}
\label{sec:limitations}



We restricted our experiments to four popular LLMs (\GPTLongName, \ClaudeLongName, \DeepSeekLongName, and \LlamaLongName) due to time and computational constraints. Given the rapid pace of model development (systems such as DeepSeek R1 \cite{guo2025deepseek}, OpenAI o3-mini \cite{o3}, and Qwen-2.5 Max \cite{qwen2024max}, it remains an open question how {\ours} would perform with these newer LLMs.

Moreover, while the user simulators fine-tuned in this study perform reasonably well, they are not perfect and may occasionally struggle to respond accurately to {\ours}.  Further details on their performance can be found in the next Appendix~\ref{sec:User_Simulator_Performance}. 

\section{User Simulator Performance}
\label{sec:User_Simulator_Performance}

\begin{table}[t!]
\centering

\resizebox{1.0\linewidth}{!}{ 
\begin{tabular}{l|l|ccc|ccc}
\toprule

\textbf{Dataset} & \textbf{LLM Model} &  
\multicolumn{3}{c|}{\makecell{\textbf{BERTScore} \\ \textbf{User (F1)} }} & 
\multicolumn{3}{c}{\makecell{\textbf{BERTScore} \\ \textbf{Overall (F1)} }} \\  
\cmidrule(lr){3-5} \cmidrule(lr){6-8}

& & 
\makecell{\textbf{Single}} & \makecell{\textbf{Multi}} & \makecell{\textbf{Both}} & 
\makecell{\textbf{Single}} & \makecell{\textbf{Multi}} & \makecell{\textbf{Both}} \\ 
\midrule

\multirow{4}{*}{\textbf{SGD}}
 & {\ours}-{\small \GPTShortName}  & 0.6879 & 0.6712 & 0.6765 & 0.6821 & 0.6761 & 0.6780 \\ 
 & {\ours}-{\small \ClaudeShortName}   & 0.7053 & 0.6972 & 0.6997 & 0.6822 & 0.6852 & 0.6843 \\ 
 & {\ours}-{\small \DeepSeekShortName}   & 0.6844 & 0.6700 & 0.6746 & 0.6600 & 0.6554 & 0.6569 \\ 
 & {\ours}-{\small \LlamaShortName}    & 0.6853 & 0.6766 & 0.6794 & 0.6365 & 0.6256 & 0.6291 \\ 
\midrule

\multirow{4}{*}{\textbf{BiTOD}} 
 & {\ours}-{\small \GPTShortName} & 0.6794 &  0.6549 & 0.6627 & 0.6889 & 0.6671 & 0.6741 \\ 
 & {\ours}-{\small \ClaudeShortName} & 0.6963 & 0.6737 & 0.6809 & 0.6841 & 0.6625 & 0.6693 \\ 
 & {\ours}-{\small \DeepSeekShortName}   & 0.6979 & 0.6502 & 0.6653 & 0.6789 & 0.6530 & 0.6612 \\ 
 & {\ours}-{\small \LlamaShortName}   & 0.6534 & 0.6370 & 0.6423 & 0.6345 & 0.5990 & 0.6104 \\ 
\bottomrule

\end{tabular}%
}
\caption{Performance of the Dialog-Level User Simulator and Entire Conversation.}
\label{table_user_scores}
\end{table}

To evaluate the performance of our user simulator, we computed BERTScore on user turns (BERTScore User) and on the concatenation of both user and system turns (BERTScore Overall). As shown in Table \ref{table_user_scores}, most BERTScores fall within the range of approximately 0.6 to 0.7, showing that the simulated user responses maintain a reasonable level of similarity to the reference.

\section{Example Dialog Responses}
\label{sec:Example_Dialog_Responses}

Tables \ref{conv_claude}, \ref{conv_gpt4o}, \ref{conv_deepseek}, and \ref{conv_llama} present an example of a multi-domain conversation where our User Simulator interacts with different variants of {\ours} (\ClaudeShortName, \GPTShortName, \DeepSeekShortName, and \LlamaShortName) to check the weather, schedule a property visit, and reserve a car. 
Additionally, Tables \ref{conv_soloist} and \ref{conv_zstod} showcase examples generated by the two baseline models SOLOIST and ZS-TOD. Both of these models struggle with task completion; SOLOIST, on the second turn, missed the API call. The same thing happened when ZS-TOD completely forgot to make the API call when handling a car reservation, ultimately failing to provide the requested information. For model errors, we use {\xmark}, and for correct API calls, we use {\cmark}.


\onecolumn

\onecolumn

\section*{Instruction Prompt Template \(P_1\)} \label{sec:prompt1}

\begin{tcolorbox}[colframe=black, colback=white, boxrule=1pt]
\textbf{\textcolor{cyan}{Task Description:}}


Your task is to generate a dialog conversation between a User and a System based on a given domain schema. I will provide a \textbf{{\textcolor{orange}{Schema}}} for \textbf{\textcolor{red!70}{\{domain\_X\}}}, which defines the structure and relevant entities, along with a corresponding dialog conversation for reference. Your goal is to analyze the relationship between the dialog and the schema and then generate a coherent and contextually appropriate dialog conversation for \textbf{\textcolor{teal}{\{domain\_Y\}}} while maintaining consistency with its schema.

\end{tcolorbox}


\begin{tcolorbox}[colframe=black, colback=white, boxrule=1pt]

Here is a \textbf{{\textcolor{orange}{Schema}}} for the \textbf{\textcolor{red!70}{\{domain\_X\}}}

\textbf{service\_name:} \textbf{\textcolor{red!70}{\{domain\_X\}}}

\textbf{\textcolor{violet}{Intents}}

\quad \textbf{{intent\_no.}} 

\quad \textbf{name}: IntentName

\quad \textbf{is\_transactional}: True/False

\quad \textbf{\textcolor{blue!70}{required\_slots}}: required slot 1, required slot 2, required slot 3, ...

\quad \textbf{\textcolor{blue!70}{optional\_slots}}: optional slot 2, optional slot 2, optional slot 3, ...

\textbf{\textcolor{orange}{Slots}}

\quad \quad \textbf{slot\_name}: slot name 1, slot name 2, slot name 3, ...

\quad \quad \textbf{possible\_values}: value 1, value 2, value 3, ...

\textit{end of schema for \textcolor{teal}{\{domain\_X\}}}

A sample \texttt{\textbf{<< Dialog Conversation >>}} between a System and a User will be fetched.
\end{tcolorbox}

\begin{tcolorbox}[colframe=black, colback=white, boxrule=1pt]
Now, understand the above conversation structure between a \underline{User} and a \underline{System}. You will be given a new \textbf{{\textcolor{orange}{Schema}}} for \textbf{\textcolor{teal}{\{domain\_Y\}}}. You have to generate a full-fledged conversation for the new domain that will be structured like the example above. 
\end{tcolorbox}

\begin{tcolorbox}[colframe=black, colback=white, boxrule=1pt]

Here is a \textbf{{\textcolor{orange}{Schema}}} for the \textbf{\textcolor{teal}{\{domain\_Y\}}}

\textbf{service\_name:} \textbf{\textcolor{teal}{\{domain\_Y\}}}

\textbf{\textcolor{violet}{Intents}}

\quad \textbf{{intent\_no.}} 

\quad \textbf{name}: IntentName

\quad \textbf{is\_transactional}: True/False

\quad \textbf{\textcolor{blue!70}{required\_slots}}: required slot 1, required slot 2, required slot 3, ...

\quad \textbf{\textcolor{blue!70}{optional\_slots}}: optional slot 2, optional slot 2, optional slot 3, ...

\textbf{\textcolor{orange}{Slots}}

\quad \quad \textbf{slot\_name}: slot name 1, slot name 2, slot name 3, ...

\quad \quad \textbf{possible\_values}: value 1, value 2, value 3, ...

\textit{end of schema for \textcolor{teal}{\{domain\_Y\}}}

\end{tcolorbox}

\begin{tcolorbox}[colframe=black, colback=white, boxrule=1pt]
Based on the above instructions and example conversation from the {domain\_X}, learn how to generate the full \underline{conversation} for the new {domain\_Y} domain.

\textit{End of Instructions.}
\end{tcolorbox}

\section*{Instruction Prompt Template \(P_2\)} \label{sec:prompt2}

\begin{tcolorbox}[colframe=black, colback=white, boxrule=1pt]
\textbf{\textcolor{cyan}{Task Description:}}

Think of yourself as an expert chat assistant specialized in the \textbf{\textcolor{red!70}{\{domain\_name\}}} domain. Your task is to generate the most natural and helpful responses for a given task-oriented dialog context. I will provide \textbf{\textcolor{orange}{Schema}} for \textbf{\textcolor{red!70}{\{domain\_name\}}}, one sample \underline{conversation} between a \underline{System} and a \underline{User}, optionally, search results from the database.  Understand the dialog relation to \textbf{\textcolor{orange}{Schema}}. You can request slot values from the \underline{User} to fulfill the \underline{User}’s current \textbf{\textcolor{violet}{intent}}. Remember that required \textbf{\textcolor{blue!70}{slots}} are more important than optional \textbf{\textcolor{blue!70}{slots}}. When making \textbf{\textbf{API}} calls, use column names from the \textbf{\textcolor{orange}{Schema}} as parameters. Match the required and optional \textbf{\textcolor{blue!70}{slots}} with the column names and use them in \textbf{\textbf{API}} calls. Before making the call, ensure you've gathered all required \textbf{\textcolor{blue!70}{slots}} from the \underline{User}. You can skip unnecessary parameters.

\end{tcolorbox}


\begin{tcolorbox}[colframe=black, colback=white, boxrule=1pt]

Here is a \textbf{{\textcolor{orange}{Schema}}} for the \textbf{\textcolor{red!70}{{domain\_name}}}

\textbf{service\_name:} \textbf{\textcolor{red!70}{\{domain\_name\}}}

\textbf{\textcolor{violet}{Intents}}

\quad \textbf{{intent\_no.}} 

\quad \textbf{name}: IntentName

\quad \textbf{is\_transactional}: True/False

\quad \textbf{\textcolor{blue!70}{required\_slots}}: required slot 1, required slot 2, required slot 3, ...

\quad \textbf{\textcolor{blue!70}{optional\_slots}}: optional slot 2, optional slot 2, optional slot 3, ...

\textbf{\textcolor{orange}{Slots}}

\quad \quad \textbf{slot\_name}: slot name 1, slot name 2, slot name 3, ...

\quad \quad \textbf{possible\_values}: value 1, value 2, value 3, ...

\textit{end of schema}

A sample \texttt{\textbf{<< Dialog Conversation >>}} between a System and a User will be fetched.
\end{tcolorbox}

\begin{tcolorbox}[colframe=black, colback=white, boxrule=1pt]
Understand the above structure of \underline{conversation} between a \underline{User} and a \underline{System}. Learn how to interact with the \underline{User} and generate the most human-like conversational response to the \underline{User}'s \textbf{\textcolor{violet}{intent}}. You may need to make \textbf{\textbf{API}} Calls and use the \textbf{\textbf{API}} Call results. Based on the above instructions and examples from the \textbf{\textcolor{red!70}{\{domain\_name\}}} domain, learn how to interact with a \underline{User} to generate the most human-like conversational response to the \underline{User}'s current \textbf{\textcolor{violet}{intent}}.

\textit{End of Instructions.}
\end{tcolorbox}

\begin{tcolorbox}[colframe=black, colback=white, boxrule=1pt]
Here are a few general \textbf{\textcolor{cyan}{Guidelines}} to follow:

- Please avoid asking for too many \textbf{\textcolor{blue!70}{slots}} in one turn; ideally, ask one slot at a time.

- Don't overwhelm the \underline{User} with too many questions or choices in one turn.

- Confirm the slot values with the \underline{User} before finalizing the \textbf{\textbf{API}} Call.

- Follow the structure of \textbf{\textbf{API}} Call from the above example whenever you are making an \textbf{\textbf{API}} Call.

- If you're unsure about something, it's always better to ask or confirm with the \underline{User}.

- Do not provide all the information in the search results to the \underline{User}. Provide details only if the \underline{User} requests them.

- If you feel the \underline{User} is confused, guide the \underline{User} with relevant suggestions and ensure it is relevant to their current \textbf{\textcolor{violet}{intent}}.

- You generate only one system response at a time and do not produce search results yourself; search results will be provided to you.

\vspace{10pt}

\textbf{Conversation history:} << conversation history >> up to turn $t$ will be fetched

\end{tcolorbox}

\begin{table*}[t!]
\centering
\caption{Multi-Domain SGD Conversation - {\ours} (\ClaudeLongName)}
\label{tab:multi_sgd_results}
\begin{tabular}{p{15cm}}
\toprule
\multicolumn{1}{c}{\textbf{SGD Multi-Domain Conversation} (\ClaudeLongName)} \\
\midrule
\multicolumn{1}{c}{\textbf{Goal:} User is trying to check the weather in \textcolor{teal}{Vancouver}, then schedule a property} \\ \multicolumn{1}{c}{visit at the \textcolor{teal}{Golf Club Manor Apartments}, and reserve a \textcolor{teal}{Hatchback} car} \\
\midrule

\textbf{User:} What's the weather going to be like in Vancouver, BC on March 2nd?\\
\midrule
\textbf{RealTOD:} I'll help you check the weather in Vancouver, BC for March 2nd. \textcolor{red}{APICall(method='GetWeather', parameters=\{ city: \textcolor{teal}{Vancouver}, date: \textcolor{teal}{2024-03-02} \})}  {\cmark}\\
\midrule
\textbf{Search Results:} \textcolor{blue}{[\{'city': 'Vancouver', 'date': '2024-03-02', 'temperature': 68, 'precipitation': 25, 'humidity': 26, 'wind': 6\}]} \\
\midrule
\textbf{RealTOD:} On March 2nd in Vancouver, the temperature will be 68 degrees Fahrenheit with a 25 percent chance of precipitation. The humidity will be at 26\% with wind speeds of 6 mph. \\
\textbf{User:} Okay, sounds good. I'd like to schedule a visit to the Golf Club Manor Apartments.\\
\textbf{RealTOD:} I can help you schedule a property visit to Golf Club Manor Apartments. I notice you were checking the weather for March 2nd. Would you like to schedule the visit for that same date? \\
\textbf{User:} Yes, I'd like to schedule the visit. \\
\midrule
\textbf{RealTOD:} \textcolor{red}{APICall(method='ScheduleVisit', parameters=\{ property\_name: \textcolor{teal}{Golf Club Manor Apartments}, visit\_date: \textcolor{teal}{2024-03-02} \})}  {\cmark} \\
\midrule
\textbf{Search Results:} \textcolor{blue}{[\{'property\_name': 'Golf Club Manor Apartments', 'address': '375 Industrial Parkway \# 314', 'area': 'Hayward', 'price': 2500000, 'phone\_number': '510\-581\-0911'\}]} \\
\midrule
\textbf{RealTOD:} I've confirmed your visit to Golf Club Manor Apartments for March 2nd. The property is located at 375 Industrial Parkway \# 314 in Hayward. Would you like to know any specific details about the property? \\
\textbf{User:} Yes, I'd like to reserve a car there. I'd like to pick it up at Indira Gandhi International Airport at 3 o'clock in the afternoon. I'd like a Hatchback. \\
\textbf{RealTOD:} I'll help you reserve a Hatchback at Indira Gandhi International Airport. Would you like to pick up the car on the same date at 3:00 PM? Also, how long will you keep the car? \\
\textbf{User:} Yes, that sounds good. I'll return the car on the 3rd of this month. \\
\textbf{RealTOD:} I'll help you reserve a Hatchback from March 2nd to March 3rd. Would you like to add insurance to your rental? \\
\textbf{User:} Yes, I'd like to add insurance. \\
\midrule
\textbf{RealTOD:} \textcolor{red}{APICall(method='ReserveCar', parameters=\{ pickup\_location: \textcolor{teal}{Indira Gandhi International Airport}, car\_type: \textcolor{teal}{Hatchback}, start\_date: \textcolor{teal}{2019-03-02}, end\_date: \textcolor{teal}{2019-03-03}, pickup\_time: \textcolor{teal}{15:00}, add\_insurance: \textcolor{teal}{True} \})}  {\cmark}\\
\midrule
\textbf{Search Results:} \textcolor{blue}{[\{'car\_name': 'Fiat Panda', 'car\_type': 'Hatchback', 'pickup\_location': 'Indira Gandhi International Airport', 'pickup\_time': '15:00', 'price\_per\_day': 39.0, 'add\_insurance': True\}]} \\
\midrule
\textbf{RealTOD:} Your car reservation has been confirmed. You'll be getting a Fiat Panda Hatchback, which will cost \$39.00 per day with insurance included. You can pick it up at Indira Gandhi International Airport on March 2nd at 3:00 PM. Would you like to know any additional details about your car rental? \\
\textbf{User:} No, thank you for your help. \\
\textbf{RealTOD:} I'm happy to assist. Have a great day! \\
\bottomrule
\end{tabular}
\label{conv_claude}
\end{table*}

\begin{table*}[ht]
\centering
\caption{Multi-SGD Conversation - {\ours} (\GPTLongName)}
\label{tab:multi_sgd_results}
\begin{tabular}{p{15cm}}
\toprule
\multicolumn{1}{c}{\textbf{SGD Multi-Domain Conversation } (\GPTLongName)} \\
\midrule
\multicolumn{1}{c}{\textbf{Goal:} User is trying to check the weather in \textcolor{teal}{Vancouver}, then schedule a property} \\ \multicolumn{1}{c}{visit at the \textcolor{teal}{Golf Club Manor Apartments}, and reserve a \textcolor{teal}{Hatchback} car} \\
\midrule

\textbf{User:} What's the weather going to be like in Vancouver, BC on March 2nd? \\
\midrule
\textbf{RealTOD:} \textcolor{red} {APICall(method='GetWeather', parameters=\{ city: \textcolor{teal}{Vancouver}, date: \textcolor{teal}{2019-03-02} \})} {\cmark}  \\
\midrule
\textbf{Search Results:} \textcolor{blue}{[\{'city': 'Vancouver', 'date': '2019-03-02', 'temperature': 67, 'precipitation': 49, 'humidity': 45, 'wind': 15\}]} \\
\midrule
\textbf{RealTOD:} The average temperature for the day will be about 67 degrees Fahrenheit, and there's a 49 percent chance of rain. \\
\textbf{User:} That sounds good. I'd like to schedule a visit to Golf Club Manor Apartments. \\
\textbf{RealTOD:} Could you confirm that you want to visit Golf Club Manor Apartments on March 2nd? \\
\textbf{User:} Sounds good. \\
\midrule
\textbf{RealTOD:} \textcolor{red}{APICall(method='ScheduleVisit', parameters=\{ property\_name: \textcolor{teal}{Golf Club Manor Apartments}, visit\_date: \textcolor{teal}{2019-03-02} \})} {\cmark}\\
\midrule
\textbf{Search Results:} \textcolor{blue}{[\{'property\_name': 'Golf Club Manor Apartments', 'address': '375 Industrial Parkway \# 314', 'number\_of\_baths': 3, 'number\_of\_beds': 3, 'phone\_number': '510-581-0911', 'price': 3950\}]} \\
\midrule
\textbf{RealTOD:} Your visit has been scheduled. \\
\textbf{User:} I would like to reserve a car there till the 3rd. \\
\textbf{RealTOD:} Do you like Hatchback, SUV, or another type of car? Would you like to pick up the car from March 2nd and from where? \\
\textbf{User:} Sounds good. Delhi Airport is a good place for me to pick it up. A hatchback sounds perfect. \\
\textbf{RealTOD:} Would you like insurance? What time will you pick it up? \\
\textbf{User:} I'd like insurance and will pick it up at 3:00 PM. \\
\textbf{RealTOD:} So you want to pick up the car from Indira Gandhi International Airport on March 2nd at 3 PM and drop it off on March 3rd? It's a Hatchback that comes with insurance. \\
\textbf{User:} Sounds perfect. \\
\midrule
\textbf{RealTOD:} \textcolor{red}{APICall(method='ReserveCar', parameters=\{ add\_insurance: \textcolor{teal}{True}, car\_type: \textcolor{teal}{Hatchback}, end\_date: \textcolor{teal}{2019-03-03}, pickup\_location: \textcolor{teal}{Indira Gandhi International Airport}, pickup\_time: \textcolor{teal}{15:00}, start\_date: \textcolor{teal}{2019-03-02} \})} {\cmark} \\
\midrule
\textbf{Search Results:} \textcolor{blue}{[\{'car\_name': 'Hyundai i20', 'car\_type': 'Hatchback', 'pickup\_location': 'Indira Gandhi International Airport', 'pickup\_time': '15:00', 'price\_per\_day': 35.00, 'add\_insurance': True\}]} \\
\midrule
\textbf{RealTOD:} Your reservation for a Hyundai i20 Hatchback has been confirmed. The cost is \$35.00 per day. If you have any more questions or need further assistance, feel free to ask! \\
\textbf{User:} Thank you. That's all I need for now. \\
\textbf{RealTOD:} Have a great day! \\
\bottomrule
\end{tabular}
\label{conv_gpt4o}
\end{table*}

\begin{table*}[ht]
\centering
\caption{Multi-SGD Conversation - {\ours} (\DeepSeekLongName)}
\label{tab:multi_sgd_results}
\begin{tabular}{p{15cm}}
\toprule
\multicolumn{1}{c}{\textbf{SGD Multi-Domain Conversation } (\DeepSeekLongName)} \\
\midrule
\multicolumn{1}{c}{\textbf{Goal:} User is trying to check the weather in \textcolor{teal}{Vancouver}, then schedule a property} \\ \multicolumn{1}{c}{visit at the \textcolor{teal}{Golf Club Manor Apartments}, and reserve a \textcolor{teal}{Hatchback} car} \\
\midrule

\textbf{User:} What's the weather going to be like in Vancouver, BC on March 2nd? \\
\midrule
\textbf{RealTOD:} \textcolor{red}{APICall(method='GetWeather', parameters=\{ city: \textcolor{teal}{Vancouver}, date: \textcolor{teal}{2019-03-02} \})} {\cmark} \\
\midrule
\textbf{Search Results:} \textcolor{blue}{[\{'city': 'Vancouver', 'date': '2019-03-02', 'temperature': 45, 'precipitation': 10, 'humidity': 65, 'wind': 12\}]} \\
\midrule
\textbf{RealTOD:} On March 2nd in Vancouver, BC, the temperature will be 45 degrees Fahrenheit with a 10 percent chance of rain. The wind speed will be 12 mph. \\
\textbf{User:} I'd like to schedule a visit to the Golf Club Manor Apartments. \\
\textbf{RealTOD:} Got it! I can help you schedule a visit to the Golf Club Manor Apartments. Could you please confirm the date you'd like to schedule the visit for? \\
\textbf{User:} I'd like to visit the property on the 2nd of this month. \\
\midrule
\textbf{RealTOD:} \textcolor{red}{APICall(method='ScheduleVisit', parameters=\{ property\_name: \textcolor{teal}{Golf Club Manor Apartments}, visit\_date: \textcolor{teal}{2019-03-02} \})}  {\cmark} \\
\midrule
\textbf{Search Results:} \textcolor{blue}{[\{'property\_name': 'Golf Club Manor Apartments', 'address': '123 Fairway Drive', 'number\_of\_baths': 2, 'number\_of\_beds': 3, 'phone\_number': '604-555-1234', 'price': 2500\}]} \\
\midrule
\textbf{RealTOD:} Your visit to the Golf Club Manor Apartments on March 2nd has been confirmed. The property is located at 123 Fairway Drive and has 3 bedrooms and 2 bathrooms. Let me know if you’d like any additional details! \\
\textbf{User:} I'd like to book a car to get there. I'd like to pick it up at Indira Gandhi International Airport, and I'd like it to be a Hatchback. \\
\textbf{RealTOD:} I'll help you book a Hatchback for your trip. Could you please provide the date and time you'd like to pick up the car, the return date, and whether you'd like to add insurance? \\
\textbf{User:} I'd like to pick up the car at 3 o'clock in the afternoon, and I'd like to return it on the 3rd of this month. I'd like to add insurance to my rental. \\
\midrule
\textbf{RealTOD:} \textcolor{red}{APICall(method='ReserveCar', parameters=\{ car\_type: \textcolor{teal}{Hatchback}, pickup\_location: \textcolor{teal}{Indira Gandhi International Airport}, start\_date: \textcolor{teal}{2019-03-02}, pickup\_time: \textcolor{teal}{15:00}, end\_date: \textcolor{teal}{2019-03-03}, add\_insurance: \textcolor{teal}{True} \})} {\cmark} \\
\midrule
\textbf{Search Results:} \textcolor{blue}{[\{'car\_name': 'Fiat Panda', 'car\_type': 'Hatchback', 'pickup\_location': 'Indira Gandhi International Airport', 'pickup\_time': '15:00', 'price\_per\_day': 39.0, 'add\_insurance': True\}]} \\
\midrule
\textbf{RealTOD:} Your car reservation has been successfully made! You'll be getting a Fiat Panda Hatchback, which will cost \$39.00 per day with insurance included. You can pick it up at Indira Gandhi International Airport on March 2nd at 3:00 PM. Let me know if you’d like any additional details! \\
\textbf{User:} Thank you for your assistance. That will be all for now. \\
\textbf{RealTOD:} I'm happy to assist. Have a great day! \\
\bottomrule
\end{tabular}
\label{conv_deepseek}
\end{table*}


\begin{table*}[ht]
\centering
\caption{Multi-SGD Conversation - {\ours} (\LlamaLongName)}
\label{tab:multi_sgd_results}
\begin{tabular}{p{15cm}}
\toprule
\multicolumn{1}{c}{\textbf{SGD Multi-Domain Conversation} (\LlamaLongName)} \\
\midrule
\multicolumn{1}{c}{\textbf{Goal:} User is trying to check the weather in \textcolor{teal}{Vancouver}, then schedule a property} \\ \multicolumn{1}{c}{visit at the \textcolor{teal}{Golf Club Manor Apartments}, and reserve a \textcolor{teal}{Hatchback} car} \\
\midrule

\textbf{User:} What's the weather going to be like in Vancouver, BC on March 2nd? \\
\textbf{RealTOD:} To provide you with the weather information for Vancouver, BC on March 2nd, I'll need to make an API call. Before I do that, I just want to confirm the details with you. You're looking for the weather in Vancouver, BC, and the specific date you're interested in is March 2nd, correct? \\
\textbf{User:} Yes, that's correct. \\
\midrule
\textbf{RealTOD:} \textcolor{red}{APICall(method='GetWeather', parameters=\{ city: \textcolor{teal}{Vancouver}, date: \textcolor{teal}{2019-03-02} \})} {\cmark}\\
\midrule
\textbf{Search Results:} \textcolor{blue}{[\{'city': 'Vancouver', 'date': '2019-03-02', 'temperature': 68, 'precipitation': 25, 'humidity': 26, 'wind': 6\}]} \\
\midrule
\textbf{RealTOD:} It seems like the search results provided multiple weather forecasts for Vancouver, BC on March 2nd. To provide a clear answer, could you specify if you're more interested in temperature, precipitation, or humidity? \\
\textbf{User:} Yes, that's what I'm looking for. I'd like to schedule a visit to the Golf Club Manor Apartments on that day. \\
\textbf{RealTOD:} To schedule a visit to the Golf Club Manor Apartments, I need to confirm the details with you. You'd like to visit on March 2nd, correct? \\
\textbf{User:} Yes, that's correct. \\
\midrule
\textbf{RealTOD:} \textcolor{red}{APICall(method='ScheduleVisit', parameters=\{ property\_name: \textcolor{teal}{Golf Club Manor Apartments}, visit\_date: \textcolor{teal}{2019-03-02} \})} {\cmark} \\
\midrule
\textbf{Search Results:} \textcolor{blue}{[\{'property\_name': 'Golf Club Manor Apartments', 'address': '375 Industrial Parkway \# 314', 'number\_of\_baths': 1, 'number\_of\_beds': 2, 'phone\_number': '510\-581\-0911', 'price': 2500000\}]} \\
\midrule
\textbf{RealTOD:} Your visit to Golf Club Manor Apartments has been scheduled. The property is located at 375 Industrial Parkway \# 314. \\
\textbf{User:} I'd like to reserve a car there and pick it up at Indira Gandhi International Airport. \\
\textbf{RealTOD:} Would you like a Sedan, Hatchback, or SUV? Also, what time would you like to pick it up? \\
\textbf{User:} I'd like a Hatchback and will pick it up at 3:00 PM. \\
\midrule
\textbf{RealTOD:} \textcolor{red}{APICall(method='ReserveCar', parameters=\{ add\_insurance: \textcolor{teal}{True}, car\_type: \textcolor{teal}{Hatchback}, end\_date: \textcolor{teal}{2019-03-03}, pickup\_location: \textcolor{teal}{Indira Gandhi International Airport}, pickup\_time: \textcolor{teal}{15:00}, start\_date: \textcolor{teal}{2019-03-02} \})} {\cmark}\\
\midrule
\textbf{Search Results:} \textcolor{blue}{[\{'car\_name': 'Fiat Panda', 'car\_type': 'Hatchback', 'pickup\_location': 'Indira Gandhi International Airport', 'pickup\_time': '15:00', 'price\_per\_day': 39.00, 'add\_insurance': True\}]} \\
\midrule
\textbf{RealTOD:} Your reservation for a Fiat Panda Hatchback has been confirmed. The cost is \$39.00 per day. \\
\textbf{User:} Thank you very much. You've been very helpful. That will be all for now. \\
\textbf{RealTOD:} Have a great day! \\
\bottomrule
\end{tabular}
\label{conv_llama}
\end{table*}

\begin{table*}[ht]
\centering
\caption{Multi-SGD Conversation Baseline Model Dialog Example -(SOLOIST)}
\label{tab:multi_sgd_results}
\begin{tabular}{p{15cm}}
\toprule
\multicolumn{1}{c}{\textbf{SGD Multi-Domain Conversation} (SOLOIST)} \\
\midrule
\multicolumn{1}{c}{\textbf{Goal:} User is trying to check the weather in \textcolor{teal}{Vancouver}, then schedule a property} \\ \multicolumn{1}{c}{visit at the \textcolor{teal}{Golf Club Manor Apartments}, and reserve a \textcolor{teal}{Hatchback} car} \\
\midrule

\textbf{User:} I'd like to find out what the weather is going to be like on 2nd of March in Vancouver, BC. \\
\midrule
\textbf{SOLOIST:} \textcolor{red}{APICall: (Missed API Call Here)} {\xmark}\\
\midrule
\textbf{Search Results:} \textcolor{blue}{[Missed Search Results]} {\xmark}\\
\midrule

\textbf{SOLOIST:} The average temperature for the day should be 67 degrees Fahrenheit with a 49 percent chance of rain. \\
\textbf{User:} That sounds good. I'd like to schedule a visit to Golf Club Manor Apartments. \\
\textbf{SOLOIST:} What time do you want to pick it up? \\
\textbf{User:} Sounds good. \\
\midrule
\textbf{SOLOIST:} \textcolor{red}{APICall(method='ScheduleVisit', parameters=\{ property\_name: \textcolor{teal}{Golf Club Manor Apartments}, visit\_date: \textcolor{teal}{2019-03-02nd} \})} {\cmark}\\
\midrule
\textbf{Search Results:} \textcolor{blue}{[\{'address': '375 Industrial Parkway \# 314', 'has\_garage': False, 'in\_unit\_laundry': True, 'number\_of\_baths': 3, 'number\_of\_beds': 3, 'phone\_number': '510-581-0911', 'price': 3950, 'property\_name': 'Golf Club Manor Apartments', 'visit\_date': '2019-03-02'\}]} \\
\midrule
\textbf{SOLOIST:} Your visit has been scheduled. \\
\textbf{User:} I would like to reserve a car there till the 3rd. \\
\textbf{SOLOIST:} Could I help you with anything else? \\
\textbf{User:} Sounds good. Delhi Airport is a good place for me to pick it up. A hatchback sounds perfect. \\
\textbf{SOLOIST:} Please confirm: You want me to reserve a standard car from YVR International Airport on March 2nd at 3:30 pm until March 11th. \\
\textbf{User:} I'd like insurance and will pick it up at 3 PM. \\
\textbf{SOLOIST:} Do you want to pick it up on March 1st? You'll be picking it up at 3:30 pm. Is that right? \\
\textbf{User:} Sounds perfect. \\
\midrule
\textbf{SOLOIST:} \textcolor{red}{APICall(method='GetCarsAvailable', parameters=\{ add\_insurance: \textcolor{teal}{True}, car\_type: Standard, end\_date: \textcolor{teal}{2019-03-03}, pickup\_location: \textcolor{teal}{YVR International Airport}, pickup\_time: \textcolor{teal}{15:30}, start\_date: \textcolor{teal}{2019-03-01} \})} {\xmark} \\

\midrule
\textbf{Search Results:} \textcolor{blue}{[\{'car\_name': 'Hyundai Elantra', 'car\_type': 'Standard', 'pickup\_location': 'VVR International Airport', 'pickup\_time': '15:30', 'price\_per\_day': 34.0, 'add\_insurance': True\}]}  \\
\midrule
\textbf{SOLOIST:} Your car has been reserved. \\
\textbf{SOLOIST:} The car has been reserved. It is a Hyundai Elantra and it is \$34. \\
\textbf{User:} What is the name of the car and how much is it per day? \\
\textbf{SOLOIST:} It is \$34.00 per day and it's called a Hyundai Elantra. \\
\textbf{User:} Thanks for helping. That's it for now. \\
\textbf{SOLOIST:} Have a great day. \\
\bottomrule
\end{tabular}
\label{conv_soloist}
\end{table*}

\begin{table*}[ht]
\centering
\caption{Multi-SGD Conversation Baseline Model Dialog Example - (ZS-TOD)}
\label{tab:multi_sgd_results}
\begin{tabular}{p{15cm}}
\toprule
\multicolumn{1}{c}{\textbf{SGD Multi-Domain Conversation} (ZS-TOD)} \\
\midrule
\multicolumn{1}{c}{\textbf{Goal:} User is trying to check the weather in \textcolor{teal}{Vancouver}, then schedule a property} \\ \multicolumn{1}{c}{visit at the \textcolor{teal}{Golf Club Manor Apartments}, and reserve a \textcolor{teal}{Hatchback} car} \\
\midrule

\textbf{User:} I'd like to find out what the weather is going to be like on 2nd of March in Vancouver, BC. \\
\midrule
\textbf{ZS-TOD:} \textcolor{red}{APICall(method='GetWeather', parameters=\{ city: \textcolor{cyan}{Vancouver}, date: \textcolor{cyan}{2019-03-02} \})} {\cmark} \\
\midrule
\textbf{Search Results:} \textcolor{blue}{[{ city: Vancouver, date: 2019-03-02, humidity: 45, precipitation: 49, temperature: 67, wind: 15 }]} \\
\midrule
\textbf{ZS-TOD:} The average temperature for the day will be about 67 degrees Fahrenheit and there's a 49 percent chance of rain. \\
\textbf{User:} That sounds good. I'd like to schedule a visit to Golf Club Manor Apartments. \\
\textbf{ZS-TOD:} Could you confirm that you want to visit Golf Club Manor Apartments on March 2nd? \\
\textbf{User:} Sounds good. \\
\midrule
\textbf{ZS-TOD:} \textcolor{red}{APICall(method='ScheduleVisit', parameters=\{ property\_name: \textcolor{cyan}{Golf Club Manor Apartments}, visit\_date: \textcolor{cyan}{2019-03-02} \})} {\cmark} \\
\midrule
\textbf{Search Results:} \textcolor{blue}{[{ address: 375 Industrial Parkway \# 314, has\_garage: False, in\_unit\_laundry: True, number\_of\_baths: 3, number\_of\_beds: 3, phone\_number: 510-581-0911, price: 3950, property\_name: Golf Club Manor Apartments, visit\_date: 2019-03-02 }]} \\
\midrule
\textbf{ZS-TOD:} Your visit has been scheduled. \\
\textbf{User:} I would like to reserve a car there till the 3rd. \\
\textbf{ZS-TOD:} Do you like Hatchback, SUV, or another type of car? Would you like to pick up the car from March 2nd and from where? \\
\textbf{User:} Sounds good. Delhi Airport is a good place for me to pick it up. A hatchback sounds perfect. \\
\textbf{ZS-TOD:} Would you like insurance? What time will you pick it up? \\
\textbf{User:} I'd like insurance and will pick it up at 3 PM. \\
\textbf{ZS-TOD:} So you want to pick up the car from Indira Gandhi International Airport on March 2nd at 3 PM and drop it off on March 3rd? It's a Hatchback that comes with insurance. \\

\midrule
\textbf{ZS-TOD:} \textcolor{red}{APICall: (Missed API Call Here)} {\xmark}\\
\midrule
\textbf{Search Results:} \textcolor{blue}{[Missed Search Results]} {\xmark}\\
\midrule

\textbf{User:} What is the name of the car and how much is it per day? \\
\textbf{ZS-TOD:} It is \$39.00 per day and it's called a Fiat Panda. \\
\textbf{User:} Thanks for helping. That's it for now. \\
\textbf{ZS-TOD:} Have a great day. \\
\bottomrule
\end{tabular}
\label{conv_zstod}
\end{table*}

\label{sec:appendix}

\onecolumn
\section{User Study Instructions}
\label{sec:templates}

\section*{Disclaimers of any risks to participants or annotators}

There are no significant risks associated with participating in this study. However, annotators may experience mild fatigue or cognitive strain due to prolonged reading and evaluation of multiple conversations. If you feel discomfort or fatigue, please take breaks as needed.

\section*{Instructions for Human Study Participants}

Your task is to evaluate model-generated responses in multi-turn, task-oriented conversations based on the following criteria:
\begin{enumerate}
    \item Fluency
    \item Informativeness
    \item Task Completion
\end{enumerate}

\section*{Task Overview}
\begin{itemize}
    \item You will be presented with multiple conversations, where a user interacts with a model to complete a task (e.g., booking a flight).
    \item Your job is to rate each model response independently using a 1-5 scale based on the provided criteria.
    \item This study is blind --- you will not know which model produced which response.
\end{itemize}

\section*{How to Rate Responses}
You will assign a score for each response based on the following:

\subsection*{1. Fluency}
Measures whether the response is grammatically correct, well-structured, and natural to read.

\begin{itemize}
    \item 1 = Unnatural, grammatically incorrect, or hard to read.
    \item 5 = Perfectly fluent, natural, and error-free.
\end{itemize}

\textbf{Example (Good Fluency)}
\begin{quote}
User: I need to book a flight to New York next Monday.\\
Model: Sure! What time would you like to depart?
\end{quote}

\textbf{Example (Poor Fluency)}
\begin{quote}
User: I need to book a flight to New York next Monday.\\
Model: Sure! You next Monday flight want time what?
\end{quote}

\subsection*{2. Informativeness}
Measures whether the response provides useful and relevant information to advance the task.

\begin{itemize}
    \item 1 = Vague, incorrect, or unhelpful.
    \item 5 = Precise, relevant, and useful.
\end{itemize}

\textbf{Example (Good Informativeness)}
\begin{quote}
User: Can you recommend a vegetarian restaurant nearby?\\
Model: Yes! Green Leaf Café is a highly rated vegetarian restaurant.
\end{quote}

\textbf{Example (Low Informativeness)}
\begin{quote}
User: Can you recommend a vegetarian restaurant nearby?\\
Model: There are many restaurants in your area.
\end{quote}

\subsection*{3. Task Completion}
Measures whether the response successfully progresses or completes the task in the conversation.

\begin{itemize}
    \item 1 = Fails to address the request.
    \item 5 = Fully completes the task.
\end{itemize}

\textbf{Example (Good Task Completion)}
\begin{quote}
User: I’d like to book a table for two at an Italian restaurant at 7 PM today.\\
Model: I found a reservation at Bella Italia for 7 PM. Would you like me to reserve it for 2 people at 7 PM today?
\end{quote}

\textbf{Example (Failed Task Completion)}
\begin{quote}
User: I’d like to book a table for two at an Italian restaurant at 7 PM.\\
Model: There are Italian restaurants in your area.
\end{quote}

\section*{Additional Guidelines}
\begin{itemize}
    \item Be objective --- rate based on quality, not personal preference.
    \item If a response is unclear or ambiguous, leave a comment.
    \item Do not try to guess which model produced the response.
\end{itemize}

\section*{Study Duration \& Submission}
\begin{itemize}
    \item The study will take approximately 10 minutes to complete.
    \item Once you have evaluated all responses, submit your ratings.
\end{itemize}

\end{document}